\def\year{2018}\relax
\documentclass[letterpaper]{article} 
\usepackage{arxiv_style}  
\usepackage{times}  
\usepackage{helvet}  
\usepackage{courier}  
\usepackage{url}  
\usepackage{graphicx}  
\usepackage{booktabs}
\usepackage{caption}
\usepackage{sub	caption}
\usepackage{color}
\usepackage{amsfonts}
\usepackage{amsmath}
\DeclareMathOperator*{\argmax}{\arg\!\max}
\DeclareMathOperator*{\argmin}{\arg\!\min}
\usepackage{algorithm}
\usepackage{algorithmic}
\usepackage[utf8]{inputenc} 

\newcommand{\citet}[1]{\citeauthor{#1} (\citeyear{#1})}

\newcommand*{\affaddr}[1]{#1}
\newcommand*{\email}[1]{\texttt{#1}}
\newcommand*{\affmark}[1][*]{\textsuperscript{#1}}

\begin{document}
\title{Alternating Optimisation and Quadrature for Robust Control}
\author{
	Supratik Paul\affmark[1], Konstantinos Chatzilygeroudis\affmark[2], Kamil Ciosek\affmark[1], \\ \bf \Large Jean-Baptiste Mouret\affmark[2], Michael A. Osborne\affmark[3], Shimon Whiteson\affmark[1]\\
	\affaddr{\affmark[1]Department of Computer Science, \affmark[3]Department of Engineering Science, University of Oxford}\\
	\affaddr{\affmark[2]Inria, Villers-lès-Nancy, France; CNRS/Université de Lorraine, Loria, UMR 7503, Vandœuvre-lès-Nancy, France}\\
	\email{\affmark[1]\{supratik.paul,kamil.ciosek,shimon.whiteson\}@cs.ox.ac.uk}\\
	\email{\affmark[2]\{konstantinos.chatzilygeroudis,jean-baptiste.mouret\}@inria.fr}\\
	\email{\affmark[3]mosb@robots.ox.ac.uk}\\
}

\maketitle
\begin{abstract}
Bayesian optimisation has been successfully applied to a variety of reinforcement learning problems. However, the traditional approach for learning optimal policies in simulators does not utilise the opportunity to improve learning by adjusting certain environment variables: state features that are unobservable and randomly determined by the environment in a physical setting but are controllable in a simulator. This paper considers the problem of finding a robust policy while taking into account the impact of environment variables. We present \emph{Alternating Optimisation and Quadrature} (ALOQ), which uses Bayesian optimisation and Bayesian quadrature to address such settings. ALOQ is robust to the presence of significant rare events, which may not be observable under random sampling, but play a substantial role in determining the optimal policy. Experimental results across different domains show that ALOQ can learn more efficiently and robustly than existing methods.
\end{abstract}

\section{Introduction}
\label{sec:intro}

A key consideration when applying \emph{reinforcement learning} (RL) to a physical setting is the risk and expense of running trials, e.g., while learning the optimal policy for a robot. Another consideration is the robustness of the learned policies. Since it is typically  infeasible to test a policy in all contexts, it is difficult to ensure it works as broadly as intended. Fortunately, policies can often be tested in a simulator that exposes key \emph{environment variables} -- state features that are unobserved and randomly determined by the environment in a physical setting but are controllable in the simulator. This paper considers how to use environment variables to help learn robust policies.

Although running trials in a simulator is cheaper and safer than running physical trials, the computational cost of each simulated trial can still be quite high. The challenge then is to develop algorithms that are sample efficient, i.e., that minimise the number of such trials. In such settings, \emph{Bayesian Optimisation} (BO) \cite{brochu2010tutorial} is a sample-efficient approach that has been successfully applied to RL in multiple domains \cite{lizotte07,martinez2007active,martinez2009bayesian,cully2015,calandra2015}.

A na\"{i}ve approach would be to randomly sample values for the environment variables in each trial, so as to estimate expected performance. However, this approach (1) often requires testing each policy in a prohibitively large number of scenarios, and (2) is not robust to \emph{significant rare events} (SREs), i.e., it fails any time there are rare events that substantially affect expected performance. For example, rare localisation errors may mean that a robot is much nearer to an obstacle than expected, increasing the risk of a collision. Since collisions are so catastrophic, avoiding them is key to maximising expected performance, even though the factors contributing to the collision occur only rarely. In such cases, the na\"{i}ve approach will not see such rare events often enough to learn an appropriate response. 

Instead, we propose a new approach called \emph{alternating optimisation and quadrature} (ALOQ) specifically aimed towards learning policies that are robust to these rare events while being as sample efficient as possible. The main idea is to \emph{actively} select the environment variables (instead of sampling them) in a simulator thanks to a \emph{Gaussian Process} (GP) that models returns as a function of \emph{both} the policy and the environment variables and then, at each time-step, to use BO and \emph{Bayesian Quadrature} (BQ) in turn to select a policy and environment setting, respectively, to evaluate. 

We apply ALOQ to a number of problems and our results demonstrate that ALOQ learns better and faster than multiple baselines. We also demonstrate that the policy learnt by ALOQ in a simulated hexapod transfers successfully to the real robot.

\section{Related Work}

\citeauthor{frank2008} (\citeyear{frank2008}) also consider the problems posed by SREs. In particular, they propose an approach based on importance sampling (IS) for efficiently evaluating policies whose expected value may be substantially affected by rare events.  While their approach is based on \emph{temporal difference} (TD) methods, we take a BO-based policy search approach. Unlike TD methods, BO is well suited to settings in which sample efficiency is paramount and/or where assumptions (e.g., the Markov property) that underlie TD methods cannot be verified. More importantly, they assume prior knowledge of the SREs, such that they can directly alter the probability of such events during policy evaluation.  By contrast, a key strength of ALOQ is that it requires only that a set of environment variables can be controlled in the simulator, without assuming any prior knowledge about whether SREs exist, or about the settings of the environment variables that might trigger them.

More recently, \citeauthor{OFFER} (\citeyear{OFFER}) also proposed an IS based algorithm, OFFER, where the setting of the environment variable is gradually changed based on observed trials. Since OFFER is a TD method, it suffers from all the disadvantages mentioned earlier. It also assumes that the environment variable only affects the initial state as otherwise it leads to unstable IS estimates.

\citeauthor{williams_santner} (\citeyear{williams_santner}) consider a problem setting they call the \textit{design of computer experiments} that is similar to our setting, but does not specifically consider SREs. Their proposed GP-based approach marginalises out the environment variable by  alternating between BO and BQ. However, unlike ALOQ, their method is based on the EI acquisition function, which makes it computationally expensive for reasons discussed in Section \ref{sec:method}, and is applicable only to discrete environment variables. We include their method as a baseline in our experiments. Our results presented in Section \ref{sec:results} show that, compared to ALOQ, their method is unsuitable for settings with SREs. Further, their method is far more computationally expensive and fails even to outperform a baseline that randomly samples the environment variable at each step.

\citeauthor{krause2011contextual} (\citeyear{krause2011contextual}) also address optimising performance in the presence of environment variables. However, they address a fundamentally different contextual bandit setting in which the learned policy conditions on the observed environment variable.

PILCO \cite{PILCO} is a model-based policy search method that achieves remarkable sample efficiency in robot control \cite{PILCO2}. PILCO superficially resembles ALOQ in its use of GPs but the key difference is that in PILCO the GP models the transition dynamics while in ALOQ it models the returns as a function of the policy and environment variable. PILCO is fundamentally ill suited to our setting. First, it assumes that the transition dynamics are Gaussian and can be learned with a few hundred observed transitions, which is often infeasible in more complex environments (i.e., it scales poorly as the dimensionality of the state/action space increases). Second, even in simple environments, PILCO will not be able to learn the transition dynamics because in our setting the environment variable is not observed in physical trials, leading to major violations of the Gaussian assumption when those environment variables can cause SREs.

Policies found in simulators are rarely optimal when deployed on the physical agent due to the reality gap that may exist due to the inability of any simulator to model reality perfectly. EPOpt \cite{EPOpt} tries to address this by finding policies that are robust to simulators with different settings of its parameters. First, multiple instances of the simulator are generated by drawing a random sample of the simulator parameter settings. Trajectories are then sampled from each of these instances and used by a batch policy optimisation algorithm (e.g. TRPO \cite{TRPO}). While ALOQ finds a risk-neutral policy, EPOpt finds a risk-averse solution based on maximising the conditional value at risk (CVaR) by feeding the policy optimisation only the sampled trajectories whose returns are lower than the CVaR. In a risk-neutral setting, EPOpt reduces to the underlying policy optimisation algorithm with trajectories randomly sampled from different instances of the simulator. This approach will not see SREs often enough to learn an appropriate response, as we demonstrate in our experiments.

\citeauthor{RARL} (\citeyear{RARL}) also suggest a method to address the problem of finding robust policies. Their method learns a policy by training in a simulator that is adversarial in nature, i.e., the simulator settings are dynamically chosen to minimise the returns of the policy. This method requires significant prior knowledge to be able to set the simulator settings such that it provides just the right amount of challenge to the policy. Furthermore, it does not consider any settings with SREs.

\begin{figure*}[t]
	\centering
	\begin{subfigure}{0.32\linewidth}
		\centering
		\includegraphics[width=1\linewidth]{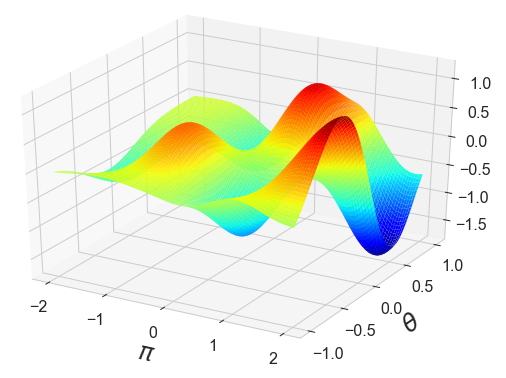}
		\caption{}
		\label{fig:GP_3D_plot}
	\end{subfigure}
	\begin{subfigure}{0.32\linewidth}
		\centering
		\includegraphics[width=1\linewidth]{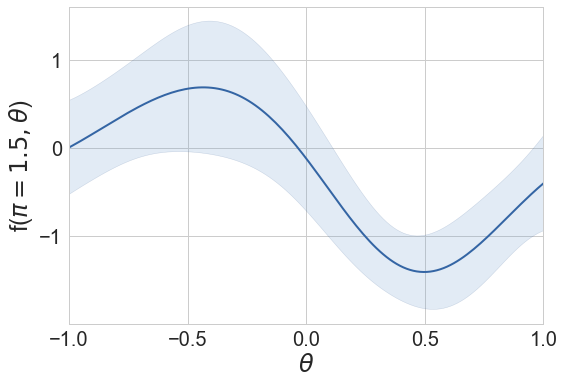}
		\caption{}
		\label{fig:pi_slice}
	\end{subfigure}
	\begin{subfigure}{0.32\linewidth}
		\centering
		\includegraphics[width=1\linewidth]{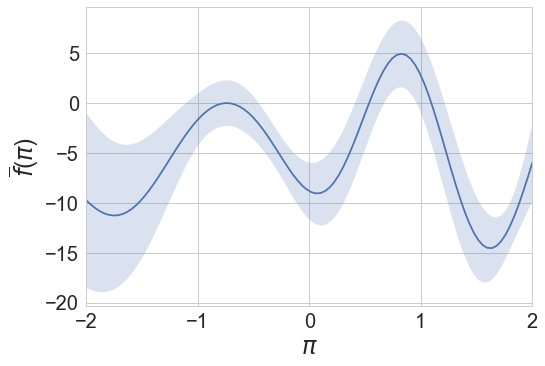}
		\caption{}
		\label{fig:BQ_plot}
	\end{subfigure}
	\caption{ALOQ models the return $f$ as a function of $(\pi,\theta)$; (a) the predicted mean based on some observed data; (b) the predicted return of $\pi=1.5$ for different $\theta$, together with the uncertainty associated with them, given $p(\theta)$; (c) ALOQ marginalises out $\theta$ and computes $\bar{f}(\pi)$ and its associated uncertainty, which is used to actively select $\pi$.}
	\label{fig:BO_BQ_explain}
\end{figure*}

\section{Background}
\label{sec:background}

GPs provide a principled way of quantifying uncertainties associated with modelling unknown functions. A GP is a distribution over functions, and is fully specified by its mean function $m(\mathbf{x})$ and covariance function $k(\mathbf{x}, \mathbf{x'})$ (see \citeauthor{GPML} (\citeyear{GPML}) for an in-depth treatment) which encode any prior belief about the nature of the function. The prior can be combined with observed values to update the belief about the function in a Bayesian way to generate a posterior distribution.

The prior mean function of the GP is often assumed to be 0 for convenience. A popular choice for the covariance function is the class of stationary functions of the form $k(\mathbf{x}, \mathbf{x'}) = k(\mathbf{x} - \mathbf{x'})$, which implies that the correlation between the function values of any two points depends only on the distance between them. 

In GP regression, it is assumed that the observed function values $\{f(\mathbf{x}_i)\}_{i=1}^{N}$ is a sample from a multivariate Gaussian distribution. The prediction for a new point $\mathbf{x}^*$ is connected with the observations through the mean and covariance functions. By conditioning on the observed data, this can be computed analytically as a Gaussian $\mathcal{N}\bigl(\mu\bigl(f(\mathbf{x}^*)\bigr), \sigma^2\bigl(f(\mathbf{x}^*)\bigr)\bigr)$:
\begin{subequations}
	\label{eq:GP_predictions}
	\begin{align}
		\mu\bigl(f(\mathbf{x}^*)\bigr) &= k(\mathbf{x}^*, \mathbf{X})(\mathbf{K}+\sigma^2_{noise} \mathbf{I})^{-1}f(\mathbf{X}) \\
		\sigma^2\bigl( f(\mathbf{x}^*)\bigr) &= k(\mathbf{x}^*,\mathbf{x}^*)\\
		&- k(\mathbf{x}^*, \mathbf{X})(\mathbf{K}+\sigma^2_{noise} \mathbf{I})^{-1}k(\mathbf{X}, \mathbf{x}^*),
	\end{align}
\end{subequations}
where $\mathbf{X}$ denotes the vector of observed inputs, $f(\mathbf{X})$ the vector of corresponding function values, and $\mathbf{K}$ is the matrix with entries $k(\mathbf{x_i}, \mathbf{x_j})$.

This property of generating estimates of the uncertainty associated with any prediction makes it particularly suited for finding the optimum of $f(\mathbf{x})$ using BO. BO requires an \emph{acquisition function} to guide the search and balance exploitation (searching the space expected to have the optimum) and exploration (searching the space which has not been explored well). Given a set of observations, the next point for evaluation is actively chosen as the $\mathbf{x}$ that maximises the acquisition function.

Two commonly used acquisition functions are \emph{expected improvement} (EI) \cite{mockus1975,jones98} and \emph{upper confidence bound} (UCB) \cite{cox92sdo,cox97sdo}. Defining $\mathbf{x}^+$ as the current optimal evaluation, i.e., $\mathbf{x}^+ = \argmax_{\mathbf{x}_i} f(\mathbf{x}_i)$, EI seeks to maximise the expected improvement over the current optimum $\alpha_{EI}(\mathbf{x}) = \mathbb{E} [I(\mathbf{x})]$, where $I(\mathbf{x}) = \max\{0,f(\mathbf{x})-f(\mathbf{x^+})\}$. By contrast, UCB does not depend on $\mathbf{x}^+$ but directly incorporates the uncertainty in the prediction by defining an upper bound: $\alpha_{UCB}(\mathbf{x}) = \mu(\mathbf{x}) + \kappa \sigma(\mathbf{x})$, where $\kappa$ controls the exploration-exploitation tradeoff.

BQ \cite{oHagan91,rasmussen2003bayesian} is a sample-efficient technique for computing integrals of the form $\bar{f} = \int f(\mathbf{x}) p(\mathbf{x}) d\mathbf{x}$, where $p(\mathbf{x})$ is a probability distribution. Using GP regression to compute the prediction for any $f(\mathbf{x})$ given some observed data, $\bar{f}$ is a Gaussian whose mean and variance can be computed analytically for particular choices of the covariance function and $p(\mathbf{x})$ \cite{briol_probabilistic_2015}. If no analytical solution exists, we can approximate the mean and variance via Monte Carlo quadrature by sampling the predictions of various $f(\mathbf{x})$.

Given some observed data $\mathcal{D}$, we can also devise acquisition functions for BQ to actively select the next point $\mathbf{x^*}$ for evaluation. A natural objective here is to select $\mathbf{x}$ that minimises the uncertainty of $\bar{f}$, i.e., $\mathbf{x^*} = \argmin_\mathbf{x} \mathbb{V}(\bar{f}|\mathcal{D}, \mathbf{x})$ \cite{osborne_duvenaud}. Due to the nature of GPs, $\mathbb{V}(\bar{f}|\mathcal{D}, \mathbf{x})$ does not depend on $f(\mathbf{x})$ and is thus computationally feasible to evaluate. \emph{Uncertainty sampling} \cite{settles_active_2010} is an alternative acquisition function that chooses the $\mathbf{x^*}$ with the maximum posterior variance: 
$\mathbf{x^*} = \argmax_\mathbf{x} \mathbb{V}(f(\mathbf{x})|\mathcal{D})$. Although simple and computationally cheap, it is not the same as reducing uncertainty about $\bar{f}$ since evaluating the point with the highest prediction uncertainty does not necessarily lead to the maximum reduction in the uncertainty of the estimate of the integral. 

Monte Carlo (MC) quadrature simply samples $(\mathbf{x}_1, \mathbf{x}_2, ... , \mathbf{x}_N)$ from $p(\mathbf{x})$ and estimates the integral as $\bar{f} \approx \frac{1}{N}\sum_{i=1}^N f(\mathbf{x}_i).$ This typically requires a large $N$ and so is less sample efficient than BQ: it should only be used if $f$ is cheap to evaluate. 
The many merits of BQ over MC, both philosophically and practically, are brought out by \citet{oHagan1987} and \citet{HenOsbGirRSPA2015}. 
Below, we will describe an active Bayesian quadrature scheme (that is, selecting points according to an acquisition function), inspired by the empirical improvements offered by those of \citet{osborne_duvenaud} and \citet{gunter14-fast-bayesian-quadrature}. 

\section{Problem Setting \& Method}
\label{sec:method}

We assume access to a computationally expensive simulator that takes as input a policy $\pi \in \mathcal{A}$ and environment variable $\theta \in \mathcal{B}$ and produces as output the return $f(\pi,\theta) \in \mathbb{R}$, where both $\mathcal{A}$ and $\mathcal{B}$ belong to some compact sets in $\mathbb{R}^{d_{\pi}}$ and $\mathbb{R}^{d_{\theta}}$, respectively. 

We also assume access to $p(\theta)$, the probability distribution over $\theta$. $p(\theta)$ may be known a priori, or it may be a posterior distribution estimated from whatever physical trials  have been conducted.  Note that we do not require a perfect simulator: any uncertainty about the dynamics of the physical world can be modelled in $p(\theta)$, i.e., some environment variables may just be simulator parameters whose correct fixed setting is not known with certainty.

Defining $f_i = f(\pi_i, \theta_i)$, we assume we have a dataset $\mathcal{D}_{1:l} = \{(\pi_1, \theta_1, f_1), (\pi_2, \theta_2, f_2), \ldots , (\pi_l, \theta_l, f_l)\}$. Our objective is to find an optimal policy $\pi^*$:
\begin{align}
	\pi^* = \argmax_\pi \bar{f}(\pi) = \argmax_\pi \mathbb{E}_\theta[f(\pi,\theta)].
\end{align}

First, consider a na\"{i}ve approach consisting of a standard application of BO that disregards $\theta$, performs BO on $\tilde{f}(\pi) = f(\pi, \theta)$ with only one input $\pi$, and attempts to estimate $\pi^*$. Formally, this approach models $\tilde{f}$ as a GP with a zero mean function and a suitable covariance function $k(\pi, \pi')$. For any given $\pi$, the variation in $f$ due to different settings of $\theta$ is treated as noise. To estimate $\pi^*$, the  na\"{i}ve approach applies BO, while sampling $\theta$ from $p(\theta)$ at each timestep. This approach will almost surely fail due to not sampling SREs often enough to learn a suitable response.

By contrast, our method ALOQ (see Alg.~\ref{alg:ALOQ}) models $f(\pi, \theta)$ as a GP: $f \sim GP(m, k)$, acknowledging both its inputs. The main idea behind ALOQ is, given $\mathcal{D}_{1:l}$, to use a BO acquisition function to select $\pi_{l+1}$ for evaluation and then use a BQ acquisition function to select $\theta_{l+1}$, conditioning on $\pi_{l+1}$.

Selecting $\pi_{l+1}$ requires maximising a BO acquisition function \eqref{eq:bo_acq} on $\bar{f}(\pi)$, which requires estimating $\bar{f}(\pi)$, together with the uncertainty associated with it. Fortunately BQ is well suited for this since it can use the GP to estimate $\bar{f}(\pi)$ together with the uncertainty associated with it. This is illustrated in Figure \ref{fig:BO_BQ_explain}.

Once $\pi_{l+1}$ is chosen, ALOQ selects $\theta_{l+1}$ by minimising a BQ acquisition function \eqref{eq:bq_acq} quantifying the uncertainty about $\bar{f}(\pi_{l+1})$. After $(\pi_{t+1}, \theta_{l+1})$ is selected, ALOQ evaluates it on the simulator and updates the GP with the new datapoint $(\pi_{l+1}, \theta_{l+1}, f_{l+1})$. Our estimate of $\pi^*$ is thus:
\begin{align}
	\label{eq:obj_GP}
	\hat{\pi}^*
	= \argmax_\pi \mathbb{E} [\bar{f}(\pi)|\mathcal{D}_{1:l+1}].
\end{align}
Although the approach described so far actively selects $\pi$ and $\theta$ through BO and BQ, it is unlikely to perform well in practice. A key observation is that the presence of SREs, which we seek to address with ALOQ, implies that the scale of $f$ varies considerably, e.g., returns in case of collision vs no collision.  This nonstationarity cannot be modelled with our stationary kernel.  Therefore, we must transform the inputs to ensure stationarity of $f$.  In particular, we employ \emph{Beta warping}, i.e., transform the inputs using Beta CDFs with parameters $(\mathbf{\alpha}, \mathbf{\beta})$ \cite{snoek2014input}. The CDF of the beta distribution on the support $0<x<1$ is given by:
\begin{equation}
\text{BetaCDF}(x,\alpha,\beta) = \int_{0}^{x} \frac{u^{\alpha-1} (1-u)^{\beta-1}}{B(\alpha,\beta)} \text{d}u,
\end{equation}
where $B(\alpha,\beta)$ is the beta function. The beta CDF is particularly suitable for our purpose as it is able to model a variety of warpings based on the settings of only two parameters $(\alpha,\beta)$. ALOQ transforms each dimension of $\pi$ and $\theta$ independently, and treats the corresponding $(\alpha,\beta)$ as hyperparameters. We assume that we are working with the transformed inputs for the rest of the paper.

While the resulting algorithm should be able to cope with SREs, the $\hat{\pi}^*$ that it returns at each iteration may still be poor, since our BQ evaluation of $\bar{f}(\pi)$ leads to a noisy approximation of the true expected return. This is particularly problematic in high dimensional settings. To address this, \emph{intensification} \cite{SPO,SPO+_Hutter}, i.e., re-evaluation of selected policies in the simulator, is essential. Therefore, ALOQ performs two simulator calls at each timestep. In the first evaluation, $(\pi_{l+1}, \theta_{l+1})$ is selected via the BO/BQ scheme described earlier. In the second stage, $(\hat{\pi}^*, \theta^*)$ is evaluated, where $\hat{\pi}^* \in \pi_{1:l+1}$ is selected using \eqref{eq:obj_GP} and $\theta^*|\hat{\pi}^*$ using the BQ acquisition function \eqref{eq:bq_acq}.

\textbf{Computing $\bar{f}(\pi)$}: For discrete $\theta$ with support $\{\theta_1, \theta_2, \ldots, \theta_{N_\theta}\}$, the estimate of the mean $\mu$ and variance $\sigma^2$ for $\bar{f}(\pi) \mid \mathcal{D}_{1:l}$ is straightforward:
\begin{subequations}
	\label{eq:predictive_fbar}
	\begin{align}
		\mu &= \frac{1}{N_\theta} \sum_{i=1}^{N_\theta} \mathbb{E}[f(\pi,\theta_i)|\mathcal{D}_{1:l}] \\
		\sigma^2 &= \frac{1}{N_\theta^2} \sum_{i=1}^{N_\theta} \sum_{j=1}^{N_\theta} Cov[f(\pi,\theta_i)|\mathcal{D}_{1:l}, f(\pi,\theta_j)|\mathcal{D}_{1:l}],
	\end{align}
\end{subequations}
where $f(\pi,\theta)$ is the prediction from the GP with mean and covariance computed using \eqref{eq:GP_predictions}. For continuous $\theta$, we apply Monte Carlo quadrature. Although this requires sampling a large number of $\theta$ and evaluating the corresponding $f(\pi,\theta) \mid \mathcal{D}_{1:l}$, it is feasible since we  evaluate $f(\pi,\theta) \mid \mathcal{D}_{1:l}$, not from the expensive simulator, but from the computationally cheaper GP.

\textbf{BO acquisition function for $\pi$}: A modified version of the UCB acquisition function is a natural choice since using \eqref{eq:predictive_fbar} we can compute it easily as
\begin{align}
	\alpha_{ALOQ}(\pi) = \mu(\bar{f}(\pi) \mid \mathcal{D}_{1:l}) + \kappa \sigma(\bar{f}(\pi) \mid \mathcal{D}_{1:l}),
	\label{eq:bo_acq}
\end{align}
and set $\pi_{l+1} = \argmax_\pi \alpha_{ALOQ}(\pi)$.

Note that although it is possible to define an EI-based acquisition function: $\alpha = \mathbb{E}_{\bar{f}(\pi)|\mathcal{D}_{1:l}} [I(\pi)]$, where $I(\pi) = \max \{ 0 , \bar{f}(\pi) - \bar{f}(\pi^+)\}$, as an alternative choice for ALOQ, it is prohibitively expensive to compute in practice. The stochastic $\bar{f}(\pi^+) \mid \mathcal{D}_{1:l}$ renders this analytically intractable. Approximating it using Monte Carlo sampling would require performing predictions on $l \times N_\theta$ points, i.e., all the $l$ observed $\pi$'s paired with all the $N_\theta$ possible settings of the environment variable, which is infeasible even for moderate $l$ as the computational complexity of GP predictions scales quadratically with the number of predictions.

\textbf{BQ acquisition function for $\theta$}: BQ can be viewed as performing policy evaluation in our approach. Since the presence of SREs leads to high variance in the returns associated with any given policy, it is of critical importance that we minimise the uncertainty associated with our estimate of the expected return of a policy. We formalise this objective through our BQ acquisition function for $\theta$: ALOQ selects $\theta_{l+1} \mid \pi_{l+1}$ by minimising the posterior variance of $\bar{f}(\pi_{l+1})$, yielding:
\begin{align}
	\theta_{l+1}|\pi_{l+1} = \argmin_\theta \mathbb{V}(\bar{f}(\pi_{l+1})|\mathcal{D}_{1:l}, \pi_{l+1}, \theta).
	\label{eq:bq_acq}
\end{align}
We also tried uncertainty sampling in our experiments. Unsurprisingly it performed worse as it is not as good at reducing the uncertainty associated with the expected return of a policy as explained in Section \ref{sec:background}.

\textbf{Properties of ALOQ}: Thanks to convergence guarantees for BO using $\alpha_{UCB}$ \cite{srinivas}, ALOQ converges if the BQ scheme on which it relies also converges.  Unfortunately, to the best of our knowledge, existing convergence guarantees \cite{Kanagawa,BQ_conv} apply only to BQ methods that do not actively select points, as \eqref{eq:bq_acq} does.
Of course, we expect such active selection to only improve the rate of convergence of our algorithms over non-active versions.
However, our empirical results in Section \ref{sec:results} show that in practice ALOQ efficiently optimises policies in the presence of SREs across a variety of tasks.

ALOQ's computational complexity is dominated by an $\mathcal{O}(l^3)$ matrix inversion, where $l$ is the sample size of the dataset $\mathcal{D}$. This cubic scaling is common to all BO methods involving GPs. The BQ integral estimation in each iteration requires only GP predictions, which are $\mathcal{O}(l^2)$.

\begin{algorithm}[H]
	\caption{ALOQ}
	\label{alg:ALOQ}
	\begin{algorithmic}[1]
		\INPUT A simulator that outputs $f = f(\pi, \theta)$, initial dataset $\mathcal{D}_{1:l}$, the maximum number of function evaluations $L$, and a GP prior.
		\FOR {$n = l+1,l+3,...,L-1$}
		\STATE Update the Beta warping parameters and transform the inputs.
		\STATE Update the GP to condition on the (transformed) dataset $\mathcal{D}_{1:l}$
		\STATE Use \eqref{eq:predictive_fbar} to estimate $p(\bar{f}|\mathcal{D}_{1:n-1})$
		\STATE Use the BO acquisition function \eqref{eq:bo_acq} to select $\pi_n = \argmax_\pi \alpha_{ALOQ}(\pi)$
		\STATE Use the BQ acquisition function \eqref{eq:bq_acq} to select $\theta_n|\pi_n = \argmin_\theta \mathbb{V}(\bar{f}(\pi_n)|\mathcal{D}_{1:n-1}, \pi_n, \theta)$
		\STATE Perform a simulator call with $(\pi_n, \theta_n)$ to obtain $f_n$ and update $\mathcal{D}_{1:n-1}$ to  $\mathcal{D}_{1:n}$
		\STATE Find $\hat{\pi}^* = \argmax_{\pi_i} \bar{f}(\pi_i) | \mathcal{D}_{1:n} \text{ and } \theta^*|\hat{\pi}^*$ using the BQ acquisition function \eqref{eq:bq_acq}.
		\STATE Perform a second simulator call with $(\hat{\pi}^*, \theta^*)$ to obtain $f_{n+1}$ and update $\mathcal{D}_{1:n}$ to $\mathcal{D}_{1:n+1}$
		\ENDFOR
		\OUTPUT $\pi^* = \argmax_{\pi_i} \bar{f}(\pi_i) \mid \mathcal{D}_{1:L} \hspace{0.5cm} i=1,2,...,L$
	\end{algorithmic}
\end{algorithm}

\begin{figure*}[t]
	\centering
	\begin{subfigure}{0.30\textwidth}
		\centering
		\includegraphics[width=1\linewidth]{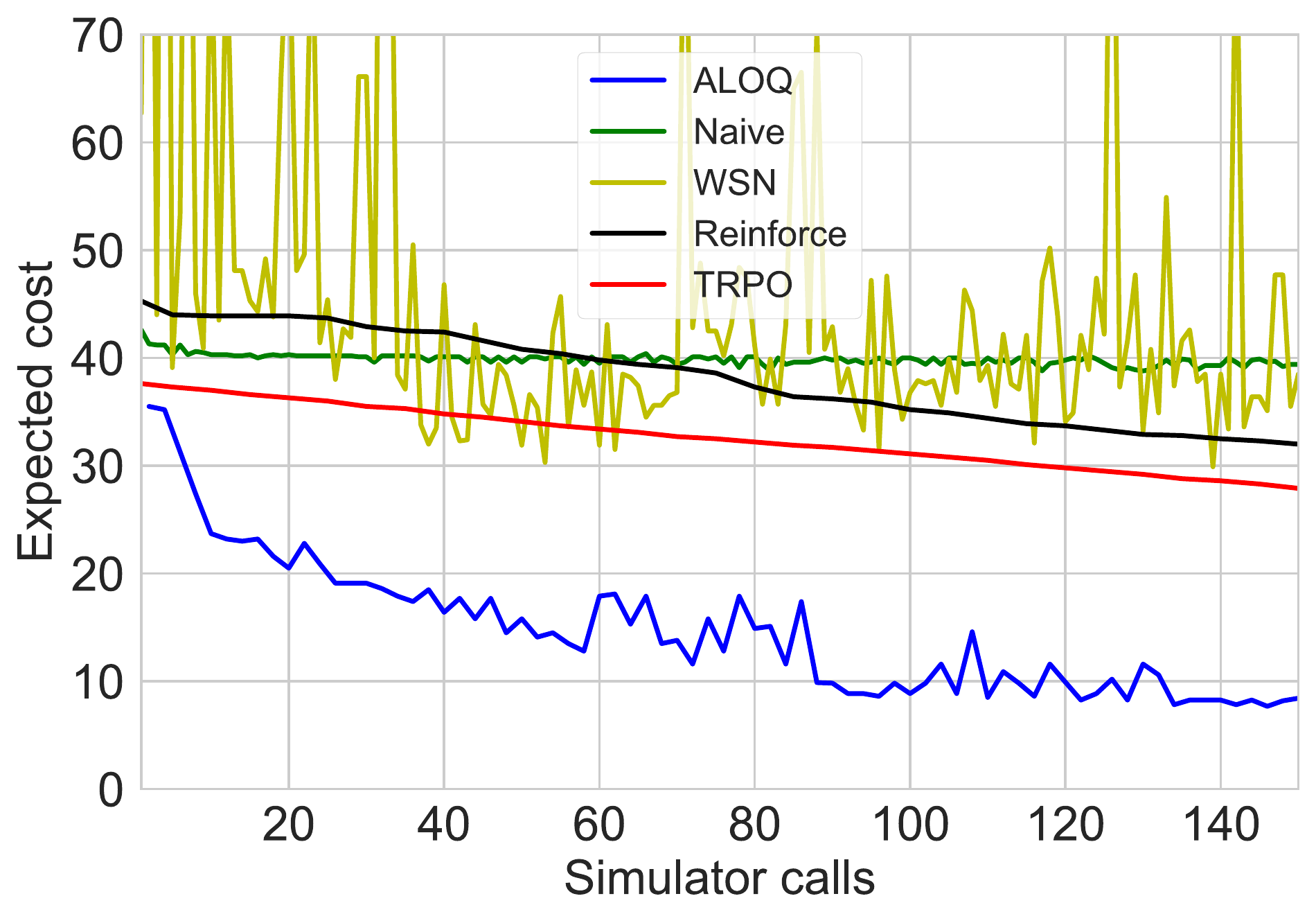}
		\caption{Expected costs of different $\hat{\pi}^*$ - Baselines}
		\label{fig:wall_bl_delta}
	\end{subfigure}\hfill
	\begin{subfigure}{0.30\textwidth}
		\centering
		\includegraphics[width=1\linewidth]{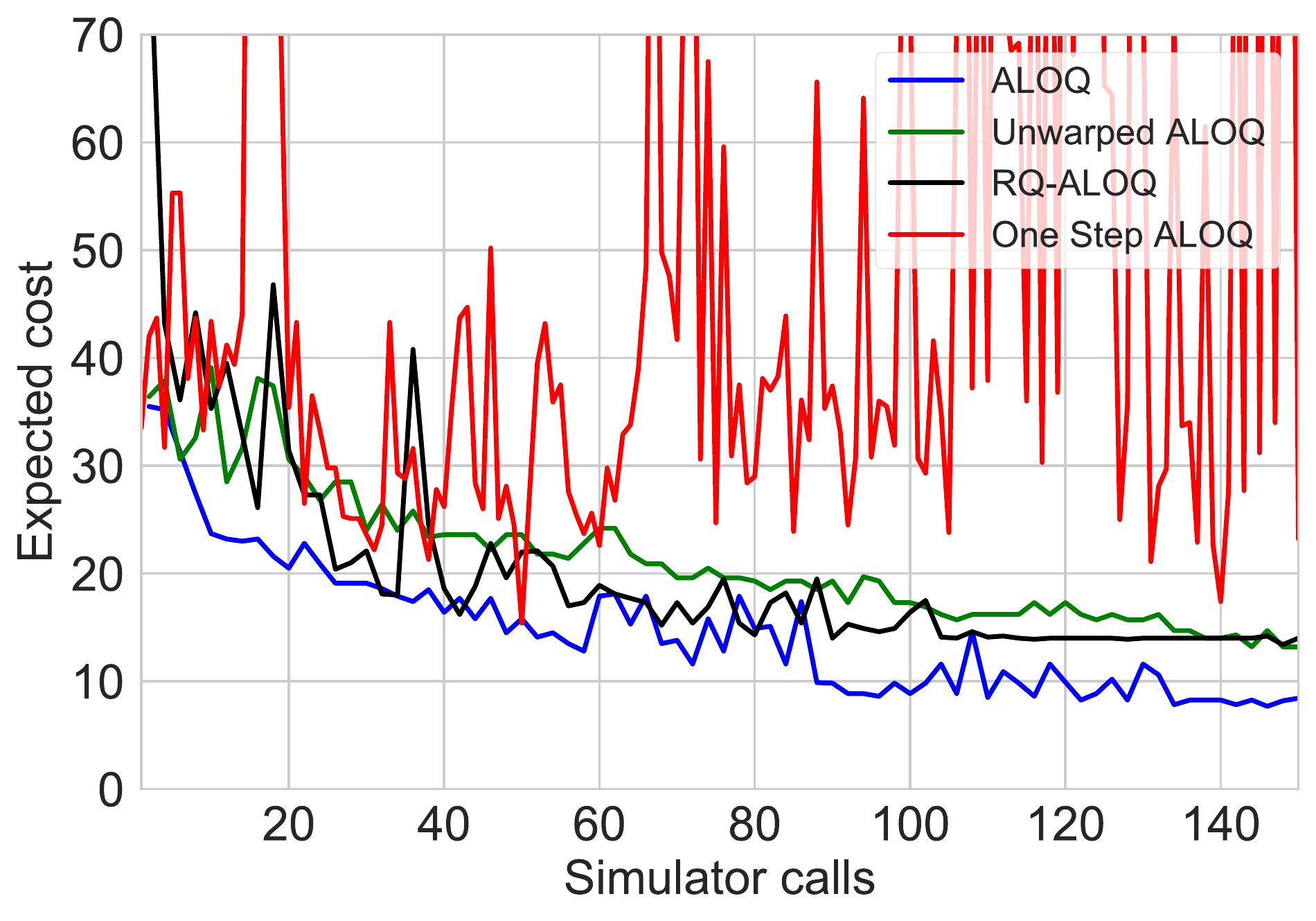}
		\caption{Expected costs of different $\hat{\pi}^*$ - Ablations}
		\label{fig:wall_abl_delta}
	\end{subfigure}\hfill
	\begin{subfigure}{0.30\textwidth}
		\centering
		\includegraphics[width=1\linewidth]{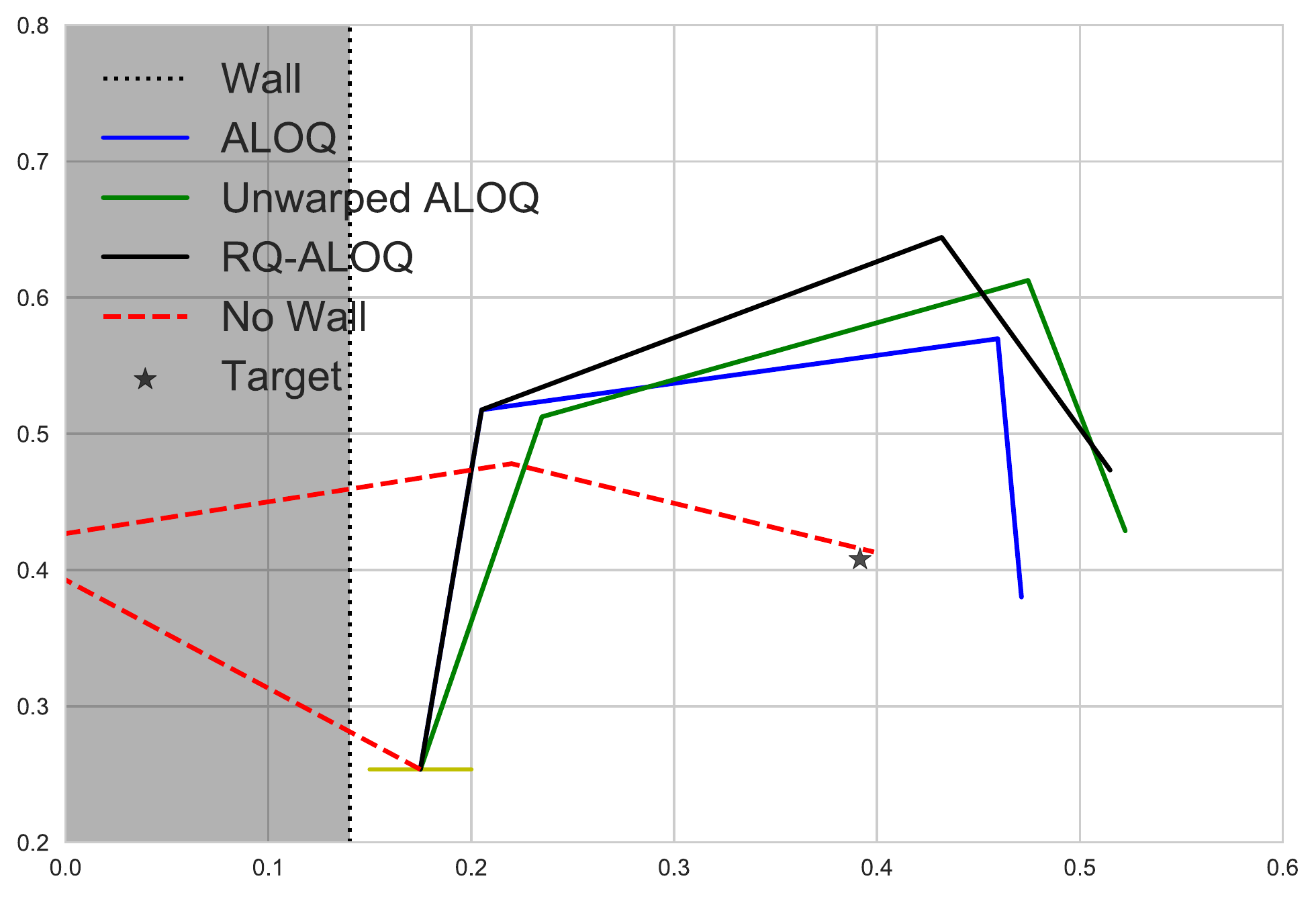}
		\caption{Learned arm configurations}
		\label{fig:wall_display}
	\end{subfigure}\hfill
	\caption{Performance and learned configurations on the robotic arm collision avoidance task.}
	\label{fig:arm_wall}
\end{figure*}

\begin{figure*}[t]
	\centering
	\begin{subfigure}{0.32\textwidth}
		\centering
		\includegraphics[width=1\linewidth]{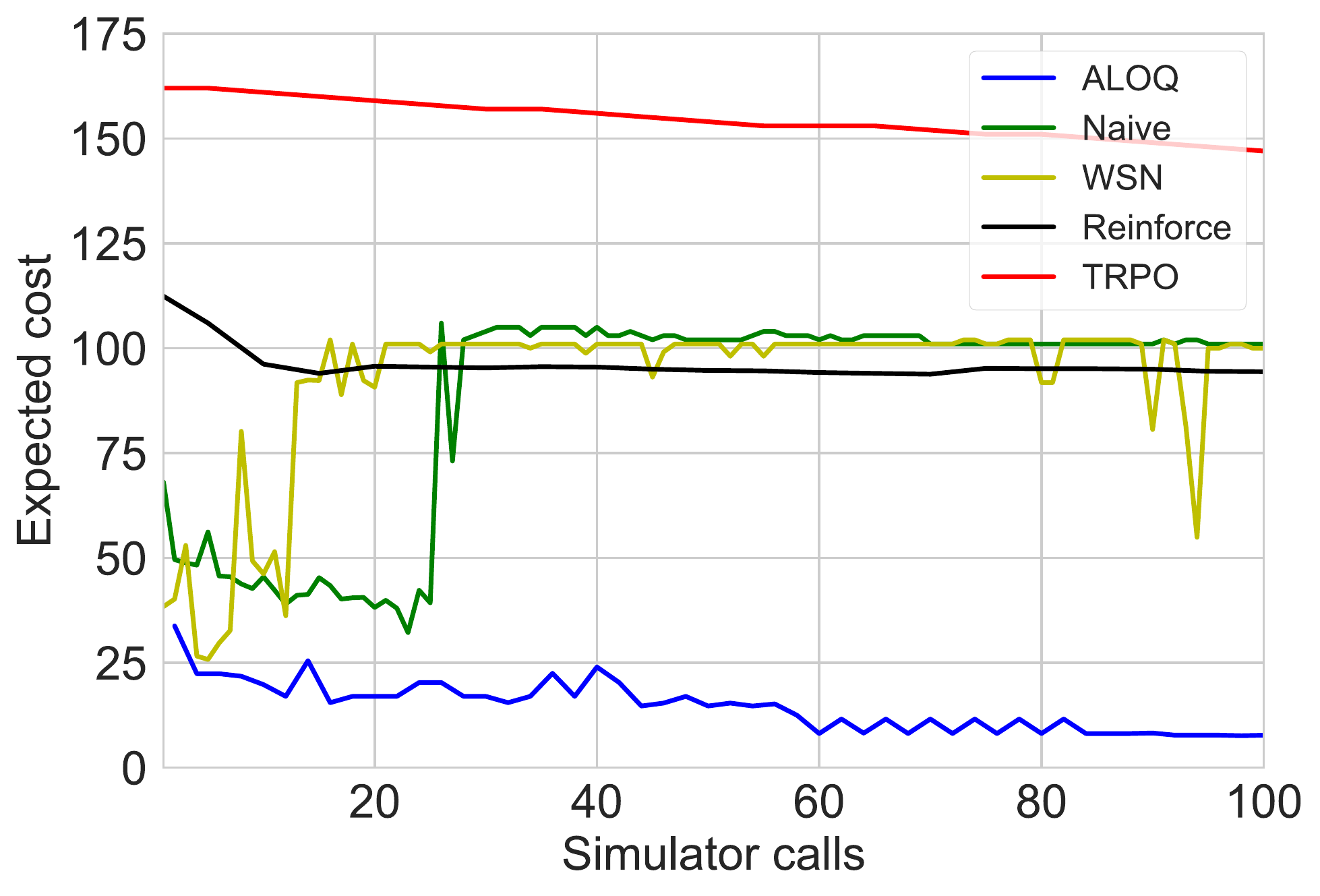}
		\caption{Expected costs of different $\hat{\pi}^*$ - Baselines}
		\label{fig:break_bl_delta}
	\end{subfigure}\hfill
	\begin{subfigure}{0.32\textwidth}
		\centering
		\includegraphics[width=1\linewidth]{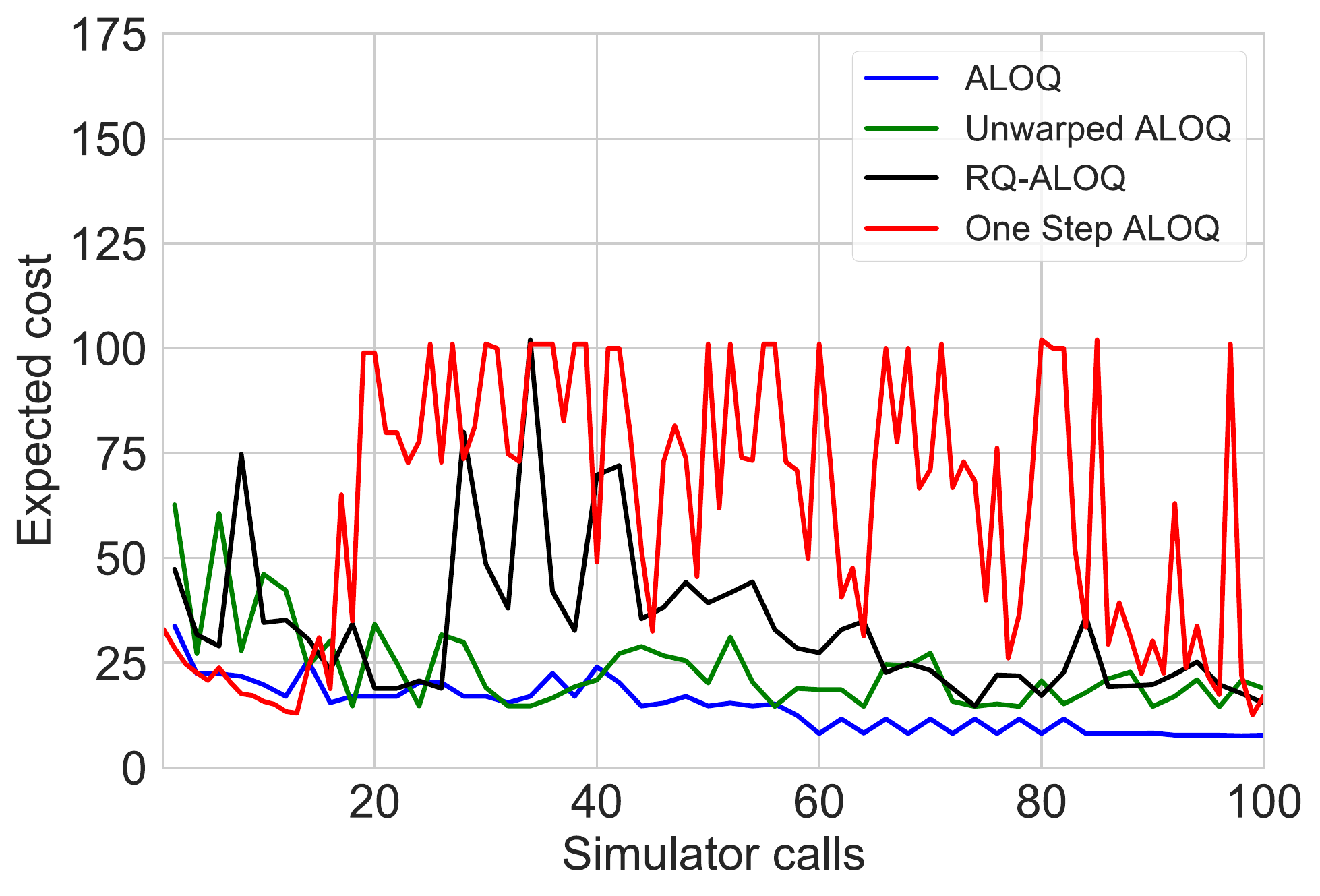}
		\caption{Expected costs of different $\hat{\pi}^*$ - Ablations}
		\label{fig:break_abl_delta}
	\end{subfigure}\hfill
	\begin{subfigure}{0.32\textwidth}
		\centering
		\includegraphics[width=1\linewidth]{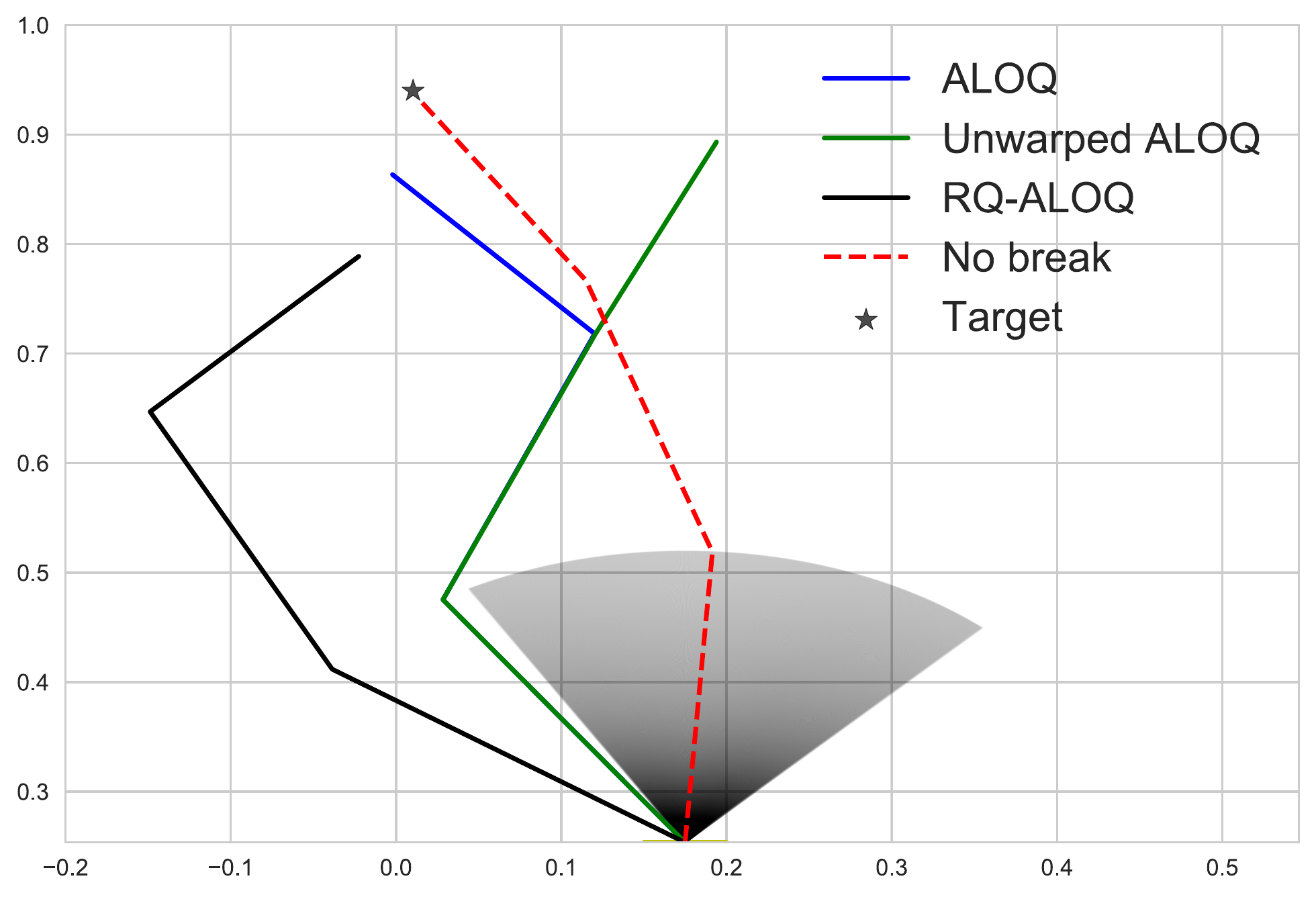}
		\caption{Learned arm configurations}
		\label{fig:break_display}
	\end{subfigure}\hfill
	\caption{Performance and learned configurations on the robotic arm joint breakage task.}
	\label{fig:arm_break}
\end{figure*}

\section{Experimental Results}
\label{sec:results}

To evaluate ALOQ we applied it to 1) a simulated robot arm control task, including a variation where $p(\theta)$ is not known a priori but must be inferred from data, and 2) a hexapod locomotion task  \cite{cully2015}. Further experiments on test functions to clearly show the how each element of ALOQ is necessary for settings with SREs is presented in the supplementary material.

We compare ALOQ to several baselines: 1) the \emph{na\"{i}ve} method described in the previous section; 2) the  method of \citeauthor{williams_santner} (\citeyear{williams_santner}), which we refer to as \emph{WSN}; 3) the simple policy gradient method Reinforce \cite{Reinforce}, and 4) the state-of-the-art policy gradient method TRPO \cite{TRPO}. To show the importance of each component of ALOQ, we also perform experiments with ablated versions of ALOQ, namely: 1) \emph{Random Quadrature ALOQ} (RQ-ALOQ), in which $\theta$ is sampled randomly from $p(\theta)$ instead of being chosen actively; 2) \emph{unwarped ALOQ}, which does not perform Beta warping of the inputs; and 3) \emph{one-step ALOQ}, which does not use intensification. All plotted results are the median of 20 independent runs. Details of the experimental setups and the variability in performance can be found in the supplementary material.

\subsection{Robotic Arm Simulator}
In this experiment, we evaluate ALOQ's performance on a robot control problem implemented in a kinematic simulator. The goal is to configure each of the three controllable joints of a robot arm such that the tip of the arm gets as close as possible to a predefined target point.

\subsubsection{Collision Avoidance}
In the first setting, we assume that the robotic arm is part of a mobile robot that has localised itself near the target. However, due to localisation errors, there is a small possibility that it is near a wall and some joint angles may lead to the arm colliding with the wall and incurring a large cost. Minimising cost entails getting as close to the target as possible while avoiding the region where the wall may be present. The environment variable in this setting is the distance to the wall.

Figures \ref{fig:wall_bl_delta} and \ref{fig:wall_abl_delta} show the expected cost (lower is better) of the arm configurations after each timestep for each method.  ALOQ, unwarped ALOQ, and RQ-ALOQ greatly outperform the other baselines.  Reinforce and TRPO, being relatively sample inefficient, exhibit a very slow rate of improvement in performance, while WSN fails to converge at all.

Figure \ref{fig:wall_display} shows the learned arm configurations, as well as the policy that would be learned by ALOQ if there was no wall (No Wall). The shaded region represents the possible locations of the wall. This plot illustrates that ALOQ learns a policy that gets closest to the target. Furthermore, while all the BO based algorithms learn to avoid the wall, active selection of $\theta$ allows ALOQ to do so more quickly: smart quadrature allows it to more efficiently observe rare events and accurately estimate their boundary. For readability we have only presented the arm configurations for algorithms which have performance comparable to ALOQ.

\subsubsection{Joint Breakage}
Next we consider a variation in which instead of uncertainty introduced by localisation, some settings of the first joint carry a 5\% probability of it breaking, which consequently incurs a large cost. Minimising cost thus entails getting as close to the target as possible, while minimising the probability of the joint breaking.

Figures \ref{fig:break_bl_delta} and \ref{fig:break_abl_delta} shows the expected cost (lower is better) of the arm configurations after each timestep for each method. Since $\theta$ is continuous in this setting, and WSN requires discrete $\theta$, it was run on a slightly different version with $\theta$ discretised by 100 equidistant points. The results are similar to the previous experiment, except that the baselines perform worse. In particular, the Na\"{i}ve baseline, WSN, and Reinforce seem to have converged to a suboptimal policy since they have not witnessed any SREs. 

Figure \ref{fig:break_display} shows the learned arm configurations together with the policy that would be learned if there were no SREs (`No break'). The shaded region represents the joint angles that can lead to failure. This figure illustrates that ALOQ learns a qualitatively different policy than the other algorithms, one which avoids the joint angles that might lead to a breakage while still getting close to the target faster than the other methods. Again for readability we only present the arm configurations for the most competitive algorithms.

\subsubsection{Performance of Reinforce and TRPO}
\begin{figure}[h]
	\centering
	\begin{subfigure}{0.85\linewidth}
		\includegraphics[width=1\linewidth]{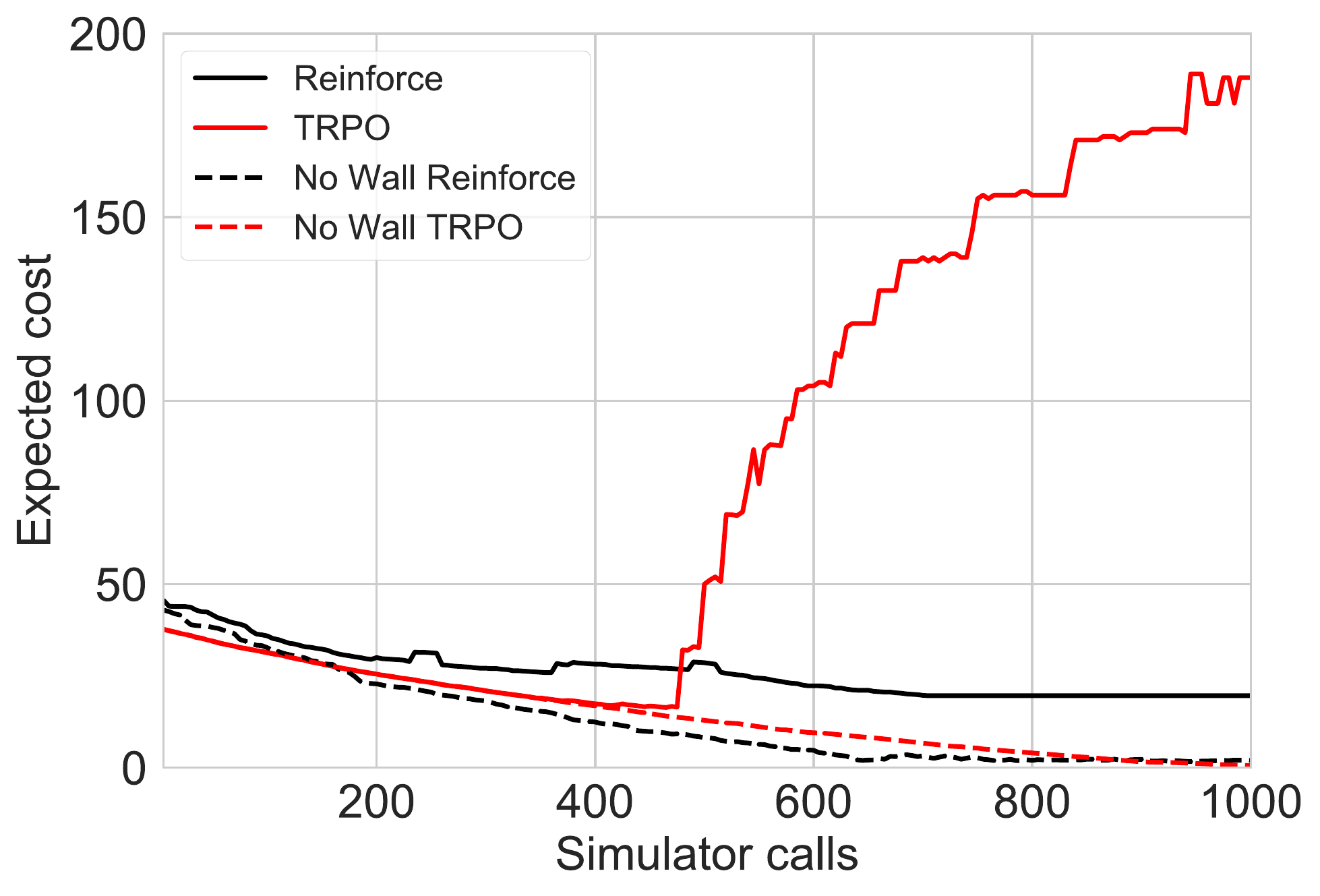}
		\caption{Collision avoidance task} 
	\end{subfigure}\hfill
	\begin{subfigure}{0.85\linewidth}
		\includegraphics[width=1\linewidth]{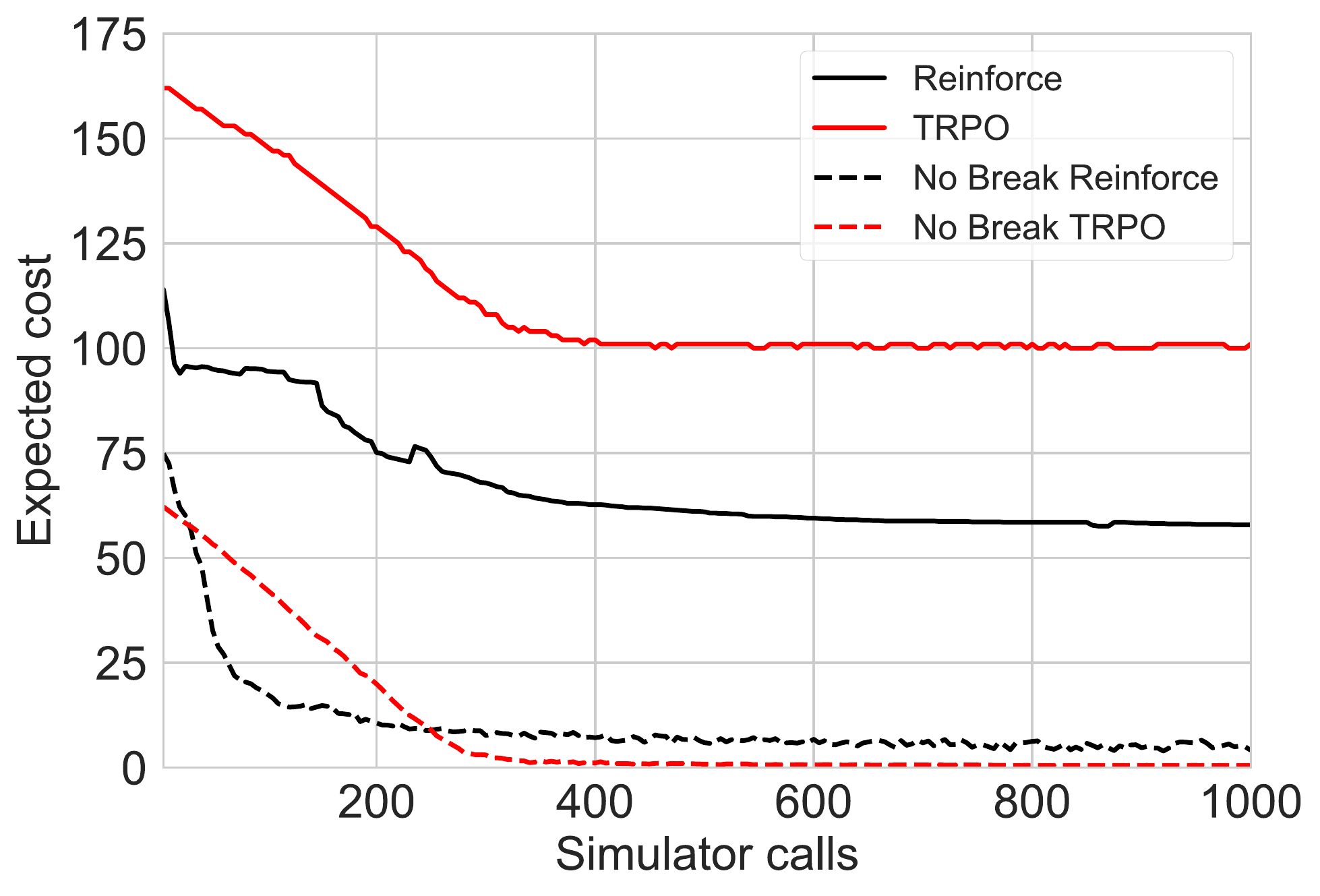}
		\caption{Arm breakage task}
	\end{subfigure}
	\caption{Performance of Reinforce and TRPO on the Robotic Arm Simulator experiments.}
	\label{fig:baselines}
\end{figure}

Both these baselines are relatively sample inefficient. However, one question that arises is whether these methods eventually find the optimal policy. To check this, we ran them for 2000 iterations with a batch size of 5 trajectories (thus a total of 10000 simulator calls). We repeated this for both the Collision Avoidance and Joint Breakage settings. The expected cost of the arm configurations after each iteration are presented in Figure \ref{fig:baselines} (we only present the results up to 1000 simulator calls for readability - there is no improvement beyond what can be seen in the plot). Both baselines can solve the tasks in settings without SREs, i.e. where there is no possibility of a collision or a breakage ('No Wall' and 'No Break' in the figures). However, in settings with SREs they converge rapidly to a suboptimal policy from which they are unable recover even if run for much longer, since they don't experience the SREs often enough. This is especially striking in the collision avoidance task where TRPO converges to a policy that has a relatively high probability of leading to a collision.

\subsubsection{Setting with unknown $p(\theta)$}
Now we consider the setting where $p(\theta)$ is not known a priori, but must be approximated using trajectories from some baseline policy. In this setting, instead of directly setting the robot arm's joint angles, we set the torque applied to each joint $(\pi)$. The final joint angles are determined by the torque and the unknown friction between the joints $(\theta)$. Setting the torque too high can lead to the joint breaking, which incurs a large cost.

We use the simulator as a proxy for both real trials as well as the simulated trials. In the first case, we simply sample $\theta$ from a uniform prior, run a baseline policy, and use the observed returns to compute an approximate posterior over $\theta$. We then use ALOQ to compute the optimal policy over this posterior (`ALOQ policy'). For comparison, we also compute the MAP of $\theta$ and the corresponding optimal policy (`MAP policy'). To show that active selection of $\theta$ is advantageous, we also compare against the policy learned by RQ-ALOQ. 

Since we are approximating the unknown $p(\theta)$ with a set of samples, it makes sense to keep the sample size relatively low for computational efficiency when finding the ALOQ policy (50 samples in this instance). However, to show that ALOQ is robust to this approximation, when comparing the performance of the ALOQ and MAP policies, we used a much larger sample size of 400 for the posterior distribution.

For evaluation, we drew 1000 samples of $\theta$ from the more granular posterior distribution and measured the returns of the three policies for each of the samples. The average cost incurred by the ALOQ policy (presented in Table \ref{tab:model_uncertainty}) was 31\% lower than that incurred by the MAP policy and 23.6\% lower than the RQ-ALOQ policy. This is because ALOQ finds a policy that slightly underperforms the MAP policy in some of  cases but avoids over 95\% of the SREs (cost $\geq$70 in Table \ref{tab:model_uncertainty}) experienced by the MAP and RQ-ALOQ policies.

\begin{table}[h]
	\caption{Comparison of the performance of ALOQ, MAP and RQ-ALOQ policies when $p(\theta)$ must be estimated}
	\label{tab:model_uncertainty}
	\centering
	\begin{tabular}{l c c c c}
		\toprule
		 & Average & \multicolumn{3}{c}{\% Episodes in Cost Range} \\
		 & Cost & 0-20 & 20-70 & $\geq$70 \\
		\midrule
		ALOQ Policy & 19.82 & 61.3\% & 38.5\% & 0.2\%  \\
		MAP Policy & 28.76 & 67.1\% & 28.7\% & 4.2\% \\
		RQ-ALOQ & 25.95 & - & 94.5\% & 5.5\% \\
		\bottomrule
	\end{tabular}
\end{table}

\subsection{Hexapod Locomotion Task}
As robots move from fully controlled environments to more complex and natural ones, they have to face the inevitable risk of getting damaged. However, it may be expensive or even impossible to decommission a robot whenever any damage condition prevents it from completing its task. Hence, it is desirable to develop methods that enable robots to recover from failure.

\emph{Intelligent trial and error} (IT\&E) \cite{cully2015} has been shown to recover from various damage conditions and thereby prevent catastrophic failure. Before deployment, IT\&E uses the simulator to create an archive of diverse and locally high performing policies for the intact robot that are  mapped to a lower dimensional \emph{behaviour space}. If the robot becomes damaged after deployment, it uses BO to quickly find the policy in the archive that has the highest performance on the damaged robot. However, it can only respond after damage has occurred.  Though it learns quickly, performance may still be poor while learning during the initial trials after damage occurs.  To mitigate this effect, we propose to use ALOQ to learn in simulation the policy with the highest expected performance across the possible damage conditions.  By deploying this policy, instead of the policy that is optimal for the intact robot, we can minimise in expectation the negative effects of damage in the period before IT\&E has learned to recover.

We consider a hexapod locomotion task with a setup similar to that of \cite{cully2015} to demonstrate this experimentally. The objective is to cross a finish line a fixed distance from its starting point. Failure  to cross the line leads to a large negative reward, while the reward for completing the task is inversely proportional to the time taken.

\begin{figure}[h]
	\centering
	\begin{subfigure}{0.85\linewidth}
		\centering
		\includegraphics[width=1\linewidth]{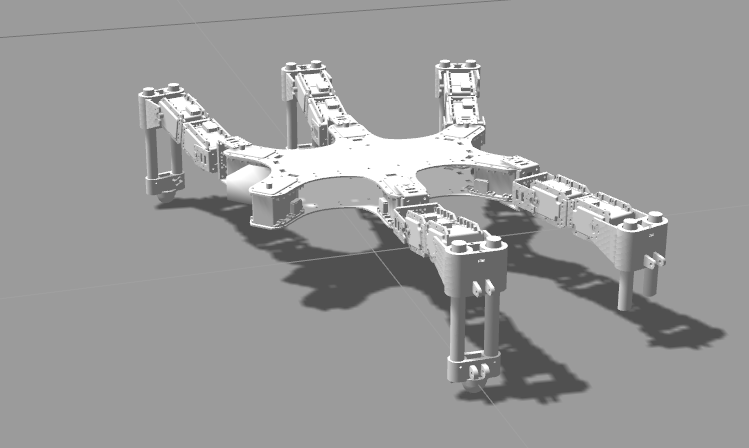}
		\caption{Hexapod with a shortened and a missing leg.}
		\label{fig:hexapod_screenshot}
	\end{subfigure}\hfill
	\begin{subfigure}{0.85\linewidth}
		\centering
		\includegraphics[width=1\linewidth]{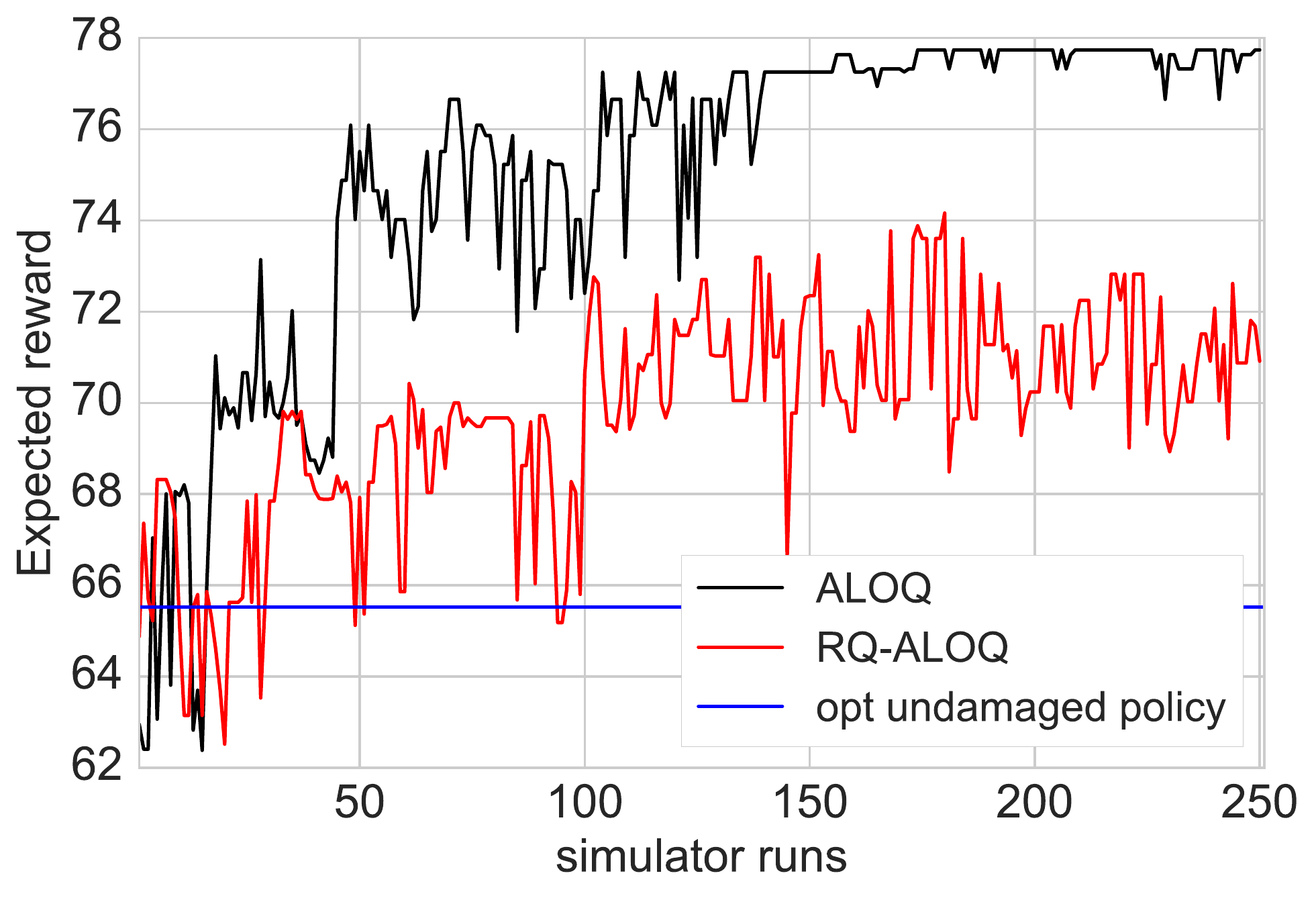}
		\caption{Expected value of $\hat{\pi}^*$}
		\label{fig:hexapod_delta}
	\end{subfigure}\hfill
	\caption{Hexapod locomotion problem.}
\end{figure}

It is possible that a subset of the legs may be damaged or broken when deployed in a physical setting. For our experiments we assume that, based on prior experience, any of the front two or back two legs can be shortened or removed with probability of 10\% and 5\% respectively, independent of the other legs, leading to 81 possible configurations. We excluded the middle two legs from our experiment as their failure had a relatively lower impact on the hexapod's movement. The configuration of the six legs acts as our environment variable. Figure \ref{fig:hexapod_screenshot} shows one such setting.

We applied ALOQ to learn the optimal policy given these damage probabilities,  but restricted the search to the policies in the archive created by \cite{cully2015}. Figure \ref{fig:hexapod_delta} shows that ALOQ finds a policy with much higher expected reward than  RQ-ALOQ. It also shows the policy that generates the maximum reward when none of the legs are damaged or broken (`opt undamaged policy'). 

To demonstrate that ALOQ learns a policy that can be applied to a physical environment, we also deployed the best ALOQ policy on the real hexapod. In order to limit the number of physical trials required to evaluate ALOQ, we limited the possibility of damage to the rear two legs. 
The learnt policy performed well on the physical robot because it optimised performance on the rare configurations that matter most for expected return (e.g., either leg shortened).

\section{Conclusions}
This paper proposed ALOQ, a novel approach to using BO and  BQ to perform sample-efficient RL in a way that is robust to the presence of significant rare events. We  empirically evaluated ALOQ on different simulated tasks involving a robotic arm simulator, and a hexapod locomotion task 
and showed how it can be also be applied to settings where the distribution of the environment variable is unknown a priori, and that it successfully transfers to a real robot. Our results demonstrated that ALOQ outperforms multiple baselines, including related methods proposed in the literature. Further, ALOQ is computationally efficient and does not require any restrictive assumptions to be made about the environment variables.

\section*{Acknowledgements}
This project has received funding from the European Research Council (ERC) under the European Union's Horizon 2020 research and innovation programme (grant agreements \#637713 and \#637972).


\bibliography{bib_ALOQ_AAAI18}
\bibliographystyle{aaai}

\clearpage
\twocolumn[
\begin{center}
	{\bf \LARGE{Supplementary Materials}
	 \vspace{0.5cm}
	}
\end{center}]

\section*{General Experimental Details}
We provide further details of our experiments in this section. 

\textbf{Covariance function:} All our experiments use a squared exponential covariance function given by:
\begin{align}
k(\mathbf{x},\mathbf{x}') = w_0 \exp (-\frac{1}{2} \sum_{d=1}^{D}(\mathbf{x}_d - \mathbf{x}'_d)^2/w_{d}^2),
\end{align}
where the hyperparameter $w_0$ specifies the variance and $\{w_i\}_{i=1}^D$ the length scales for the $D$ dimensions.

\textbf{Treatment of hyperparameters:} Instead of maximising the likelihood of the hyperparameters, we follow a full Bayesian approach and compute the marginalised posterior distribution $p(f\mid \mathcal{D})$ by first placing a hyperprior distribution on $\zeta$, the set of all hyperparameters, and then marginalising it out from $p(f\mid \mathcal{D},\zeta)$. In practice, an analytical solution for this is unlikely to exist so we estimate $\int p(f \mid \mathcal{D}, \zeta) p(\zeta \mid \mathcal{D}) d\zeta$ using Monte Carlo quadrature. Slice sampling \cite{slicesampling} was used to draw random samples from $p(\zeta \mid \mathcal{D})$.

\textbf{Choice of hyperpriors:} We assume a log-normal hyperprior distribution for all the above hyperparameters. For the variance we use $(\mu = 0, \sigma = 1)$, while for the lengthscales we use $(\mu = 0, \sigma = 0.75)$. For $\{(\alpha_i, \beta_i)\}$ we used $(\mu=0, \sigma = 0.5)$.


\textbf{Optimising the BO/BQ acquisition functions:} We used DIRECT \cite{DIRECT} to maximise the BO acquisition function $\alpha_{ALOQ}$. To minimise the BQ acquisition function, we exhaustively computed $\mathbb{V}(\bar{f}(\pi_{t+1})|\mathcal{D}_{1:t}, \pi_{t+1}, \theta)$ for each $\theta$ since this was computationally very cheap.

\section*{Robotic Arm Simulator}
The configuration of the robot arm is determined by three joint angles, each of which is normalised to lie in $[0,1]$. The arm has a reach of $[-0.54, 0.89]$ on the $x$-axis. We set $\kappa = 1.5$ for all three experiments in this section.

\subsubsection*{Collision Avoidance}
In this experiment, the target was set to the final position of the end effector for $\pi' =[0.25, 0.75, 0.8]$. The location of the wall, $\theta$, was discrete with 20 support points logarithmically distributed in $[-0.2, 0.14]$. The probability mass was distributed amongst these points such that there was only a 12\% chance of collision for $\pi'$. 

\subsubsection*{Joint Breakage}
The target for the arm breakage experiment was set to the final position of the end effector for $\pi' =[0.4, 0.2, 0.6]$. Angles between $[0.3, 0.7]$ for the first joint have an associated 5\% probability of breakage.  



\subsubsection*{Comparison of runtimes}
\begin{figure}[h]
	\begin{subfigure}{0.85\linewidth}
		\centering
		\includegraphics[width=1\linewidth]{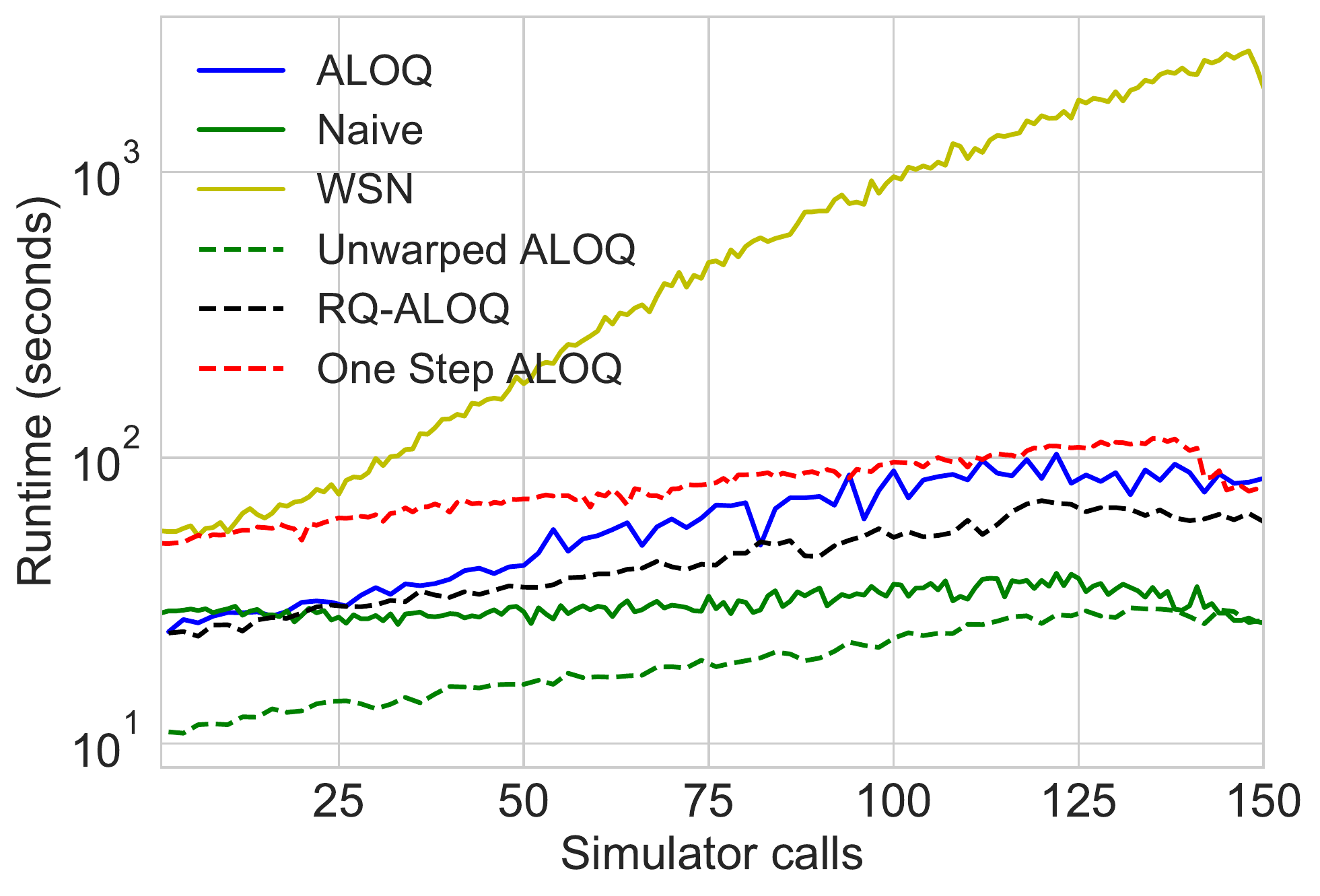}
		\caption{Collision Avoidance experiment}
		\label{fig:wall_time}
	\end{subfigure}
	\begin{subfigure}{0.85\linewidth}
		\centering
		\includegraphics[width=1\linewidth]{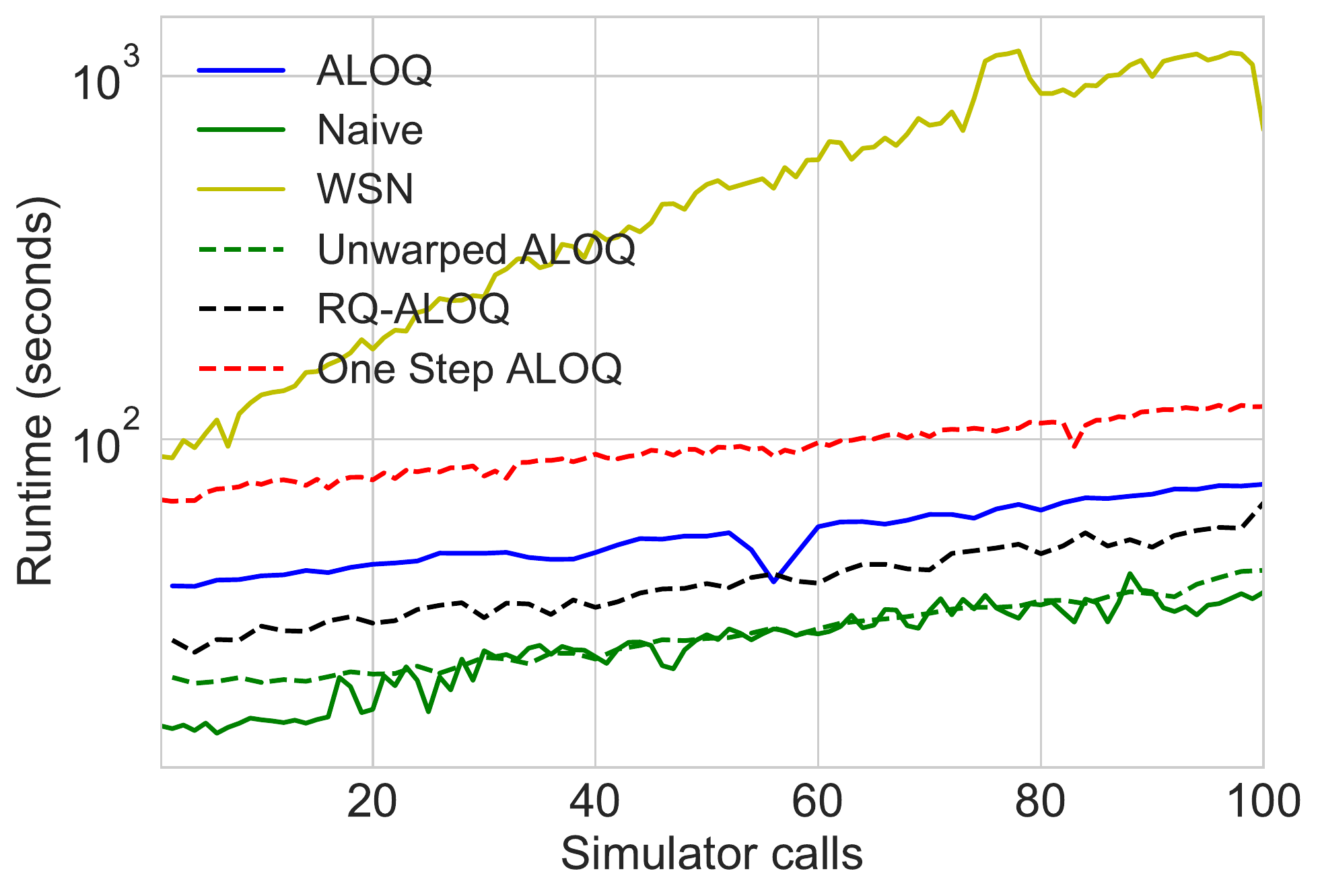}
		\caption{Joint Breakage experiment}
		\label{fig:break_time}
	\end{subfigure}\hfill
	\caption{Per-step runtime for each method on the Robotic Arm Simulator experiments}
	\label{fig:arm_time}
\end{figure}

A comparison of the per-step runtimes for the GP based methods are presented in Figure \ref{fig:arm_time}. As expected, ALOQ is once again much faster than WSN. 

\subsubsection*{Variation in performance}
The quartiles of the expected cost of the final $\hat{\pi}^*$ by each algorithm across the 20 independent runs are presented in Table \ref{tab:wall_exp}.

\begin{table}[h]
	\centering
	\caption{Quartiles of the expected cost of the final $\hat{\pi}^*$ estimated by each algorithm across 20 independent runs for the Robotic Arm Simulator experiments.}
	\begin{subtable}{1\linewidth}
		\centering
		\caption{Collision Avoidance experiment}
		\label{tab:wall_exp}
		\begin{tabular}{l c  c  c}
			\toprule
			Algorithm & Q1 & Median & Q2 \\
			\midrule
			ALOQ & 7.6 & 8.9 & 22.3 \\
			\midrule
			Na\"{i}ve & 26.7 & 40.0 & 42.1 \\
			WSN & 28.3 & 36.8 & 65.2 \\
			Reinforce & 22.0 & 32.3 & 41.5 \\
			TRPO & 27.8 & 28.3 & 28.6 \\
			\midrule
			Unwarped ALOQ & 13.6 & 17.3 & 21.0 \\
			RQ-ALOQ & 12.8 & 16.4 & 25.1 \\
			One Step ALOQ & 13.7 & 74.1 & 221.9 \\
			\bottomrule
		\end{tabular}
	\end{subtable}
	\begin{subtable}{1\linewidth}
		\centering
		\caption{Joint Breakage experiment}
		\label{tab:break_exp}
		\begin{tabular}{l c  c  c}
			\toprule
			Algorithm & Q1 & Median & Q2 \\
			\midrule
			ALOQ & 4.6 & 7.7 & 16.7 \\
			\midrule
			Na\"{i}ve & 13.2 & 100.6 & 106.7 \\
			WSN & 26.3	 & 100.5 & 103.2 \\		
			Reinforce & 61.7 & 94.5 & 97.2 \\
			TRPO & 146.0 & 148.0 & 150.0 \\
			\midrule
			Unwarped ALOQ & 5.8 & 18.9 & 34.0 \\
			RQ-ALOQ & 6.6 & 15.4 & 102.7 \\
			One Step ALOQ & 8.0 & 17.1 & 110.0 \\
			\bottomrule
		\end{tabular}
	\end{subtable}
\end{table}

\subsubsection*{Setting with unknown $p(\theta)$}
As described in the paper, in this setting we assume that $\pi \in [0,1]^3$ is the torque applied to the joints, and $\theta \in [0.5,1]$ controls the rigidity of the joints. The final joint angle is determined as $\pi/\theta$. If the torque applied to any of the joints is greater than the rigidity, (i.e. any of the angles end up $>1$), then the joint is damaged, incurring a large cost.

To simulate a set of $n$ physical trials with a baseline policy $\pi_b$, we sample $\theta$ from $U(0.5,1)$ and observe the return $f(\pi_b, \theta)$ and add iid Gaussian noise to them. The posterior can then be computed as $p(\theta|\mathcal{D}_{1:n}^b, \pi_b) \propto p(\theta)p(\mathcal{D}_{1:n}^b|\pi_b, \theta)$, where $\mathcal{D}_{1:n}^b = \{(\pi_b, f_1), (\pi_b, f_2), ..., (\pi_b, f_n) \}$. We can approximate this using slice sampling since both the prior and the likelihood are analytical.

An alternative formulation would be to corrupt the joint angles with Gaussian noise instead of the observed returns. The posterior can still be computed in this case, but instead of using slice sampling, we would have to make use of \emph{approximate Bayesian computation} \cite{rubin1984,Tavare1997,Pritchard1999}, which would be computationally expensive.

To ensure that only the information gained about $\theta$ gets carried over from the physical trials to the final ALOQ/MAP policy being learned, the target for the baseline policy was different to the target for the final policy.

As mentioned in the paper, to find the optimal policy using ALOQ, we approximated the posterior with 50 samples using a slice sampler. However, for evaluation and comparison with the MAP policy, we used a much more granular approximation with 400 samples.

\section*{Hexapod Locomotion Task}
The robot has six legs with three degrees of freedom each. We built a fairly accurate model of the robot which involved creating a URDF model with dynamic properties of each of the legs and the body, including their weights, and used the DART simulator for the dynamic physics simulation.\footnote{\url{https://dartsim.github.io}} We also used velocity actuators.

The low-level controller (or \emph{policy}) is the same open-loop controller as in \cite{cully_evolving_2015} and \cite{cully2015}. The position of the first two joints of each of the six legs is controlled by a periodic function with three parameters: an offset, a phase shift, and an amplitude (we keep the frequency fixed). The position of the third joint of each leg is the opposite of the position of the second one, so that the last segment always stays vertical. This results in 36 parameters.

The archive of policies in the behaviour space was created using the MAP-Elites algorithm \cite{mouret_illuminating_2015}. MAP-Elites searches for the highest-performing solution for each point in the duty factor space \cite{cully2015}, i.e., the time each tip of the leg spent touching the ground. MAP-Elites also acts as a dimensionality reduction algorithm and maps the high dimensional controller/policy space (in our case 36D) to the lower dimensional behaviour space (in our case 6D). We also used this lower dimensional representation of the policies in the archive as the policy search space ($\pi$) for ALOQ.

For our experiment, we set the reward such that failure to cross the finish line within 5 seconds yields zero reward, while crossing the finish line gives a reward of $100+50v$ where $v$ is the average velocity in m/s.

\section*{Further experiments}
In this section, we present the results of further experiments performed on test functions to demonstrate that each element of ALOQ is necessary for settings with SREs.

We begin with modified versions of the \emph{Branin} and \emph{Hartmann 6} test functions used by \citeauthor{williams_santner}. The modified Branin test function is a four-dimensional problem, with two  dimensions treated as discrete environment variables with a total of 12 support points, while the modified Hartmann 6 test function is six-dimensional with two dimensions treated as environment variables with a total of 49 support points. See \cite{williams_santner} for the mathematical formulation of these test functions.

\begin{figure}[t]
	\begin{subfigure}{0.85\linewidth}
		\centering
		\includegraphics[width=1\linewidth]{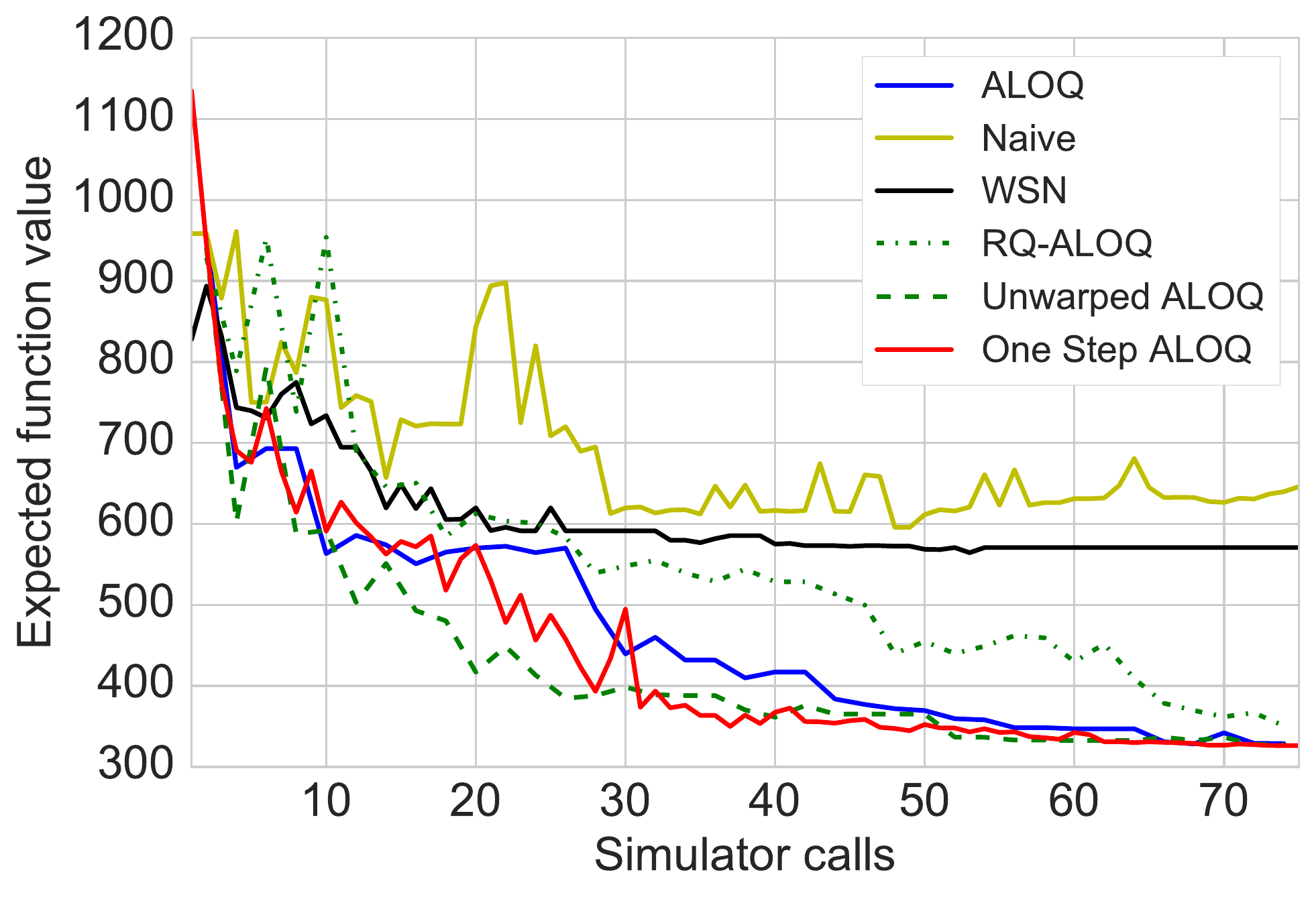}
		\caption{Branin (min) - expected value of $\hat{\pi}^*$ (lower is better)}
		\label{fig:branin_delta}
	\end{subfigure}
	\hskip 3em
	\begin{subfigure}{0.85\linewidth}
		\centering
		\includegraphics[width=1\linewidth]{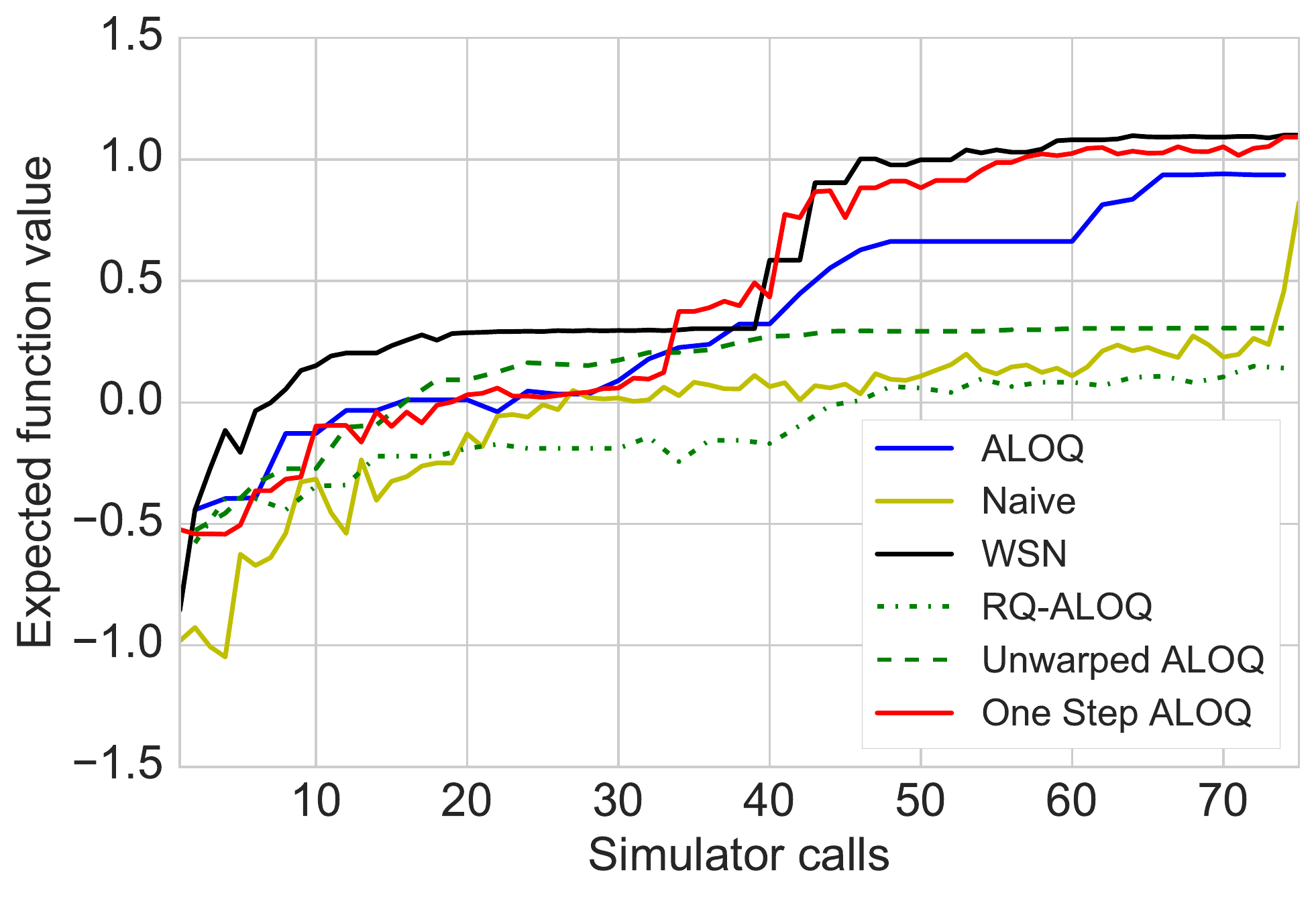}
		\caption{Hartmann 6 (max) - expected value of $\hat{\pi}^*$ (higher is better)}
		\label{fig:hart_delta}
	\end{subfigure}\hfill
	\caption{Comparison of performance of all methods on the modified Branin and Hartmann 6 test functions used by \citeauthor{williams_santner}.}
	\label{fig:wsn_perf_plots}
	\vspace{-3mm}
\end{figure}

The performance of the algorithms on the two functions is presented in Figure \ref{fig:wsn_perf_plots}. In the Branin function, ALOQ, RQ-ALOQ, unwarped ALOQ, and one-step ALOQ all substantially outperform WSN. WSN performs better in the Hartmann 6 function  as it does not get stuck in a local maximum. However, it still cannot outperform one-step ALOQ. Note that ALOQ slightly underperforms one-step ALOQ. This is not surprising: since the problem does not have SREs, the intensification procedure used by ALOQ does not yield any significant benefit.

\begin{figure}[t]
	\begin{subfigure}{0.85\linewidth}
		\centering
		\includegraphics[width=1\linewidth]{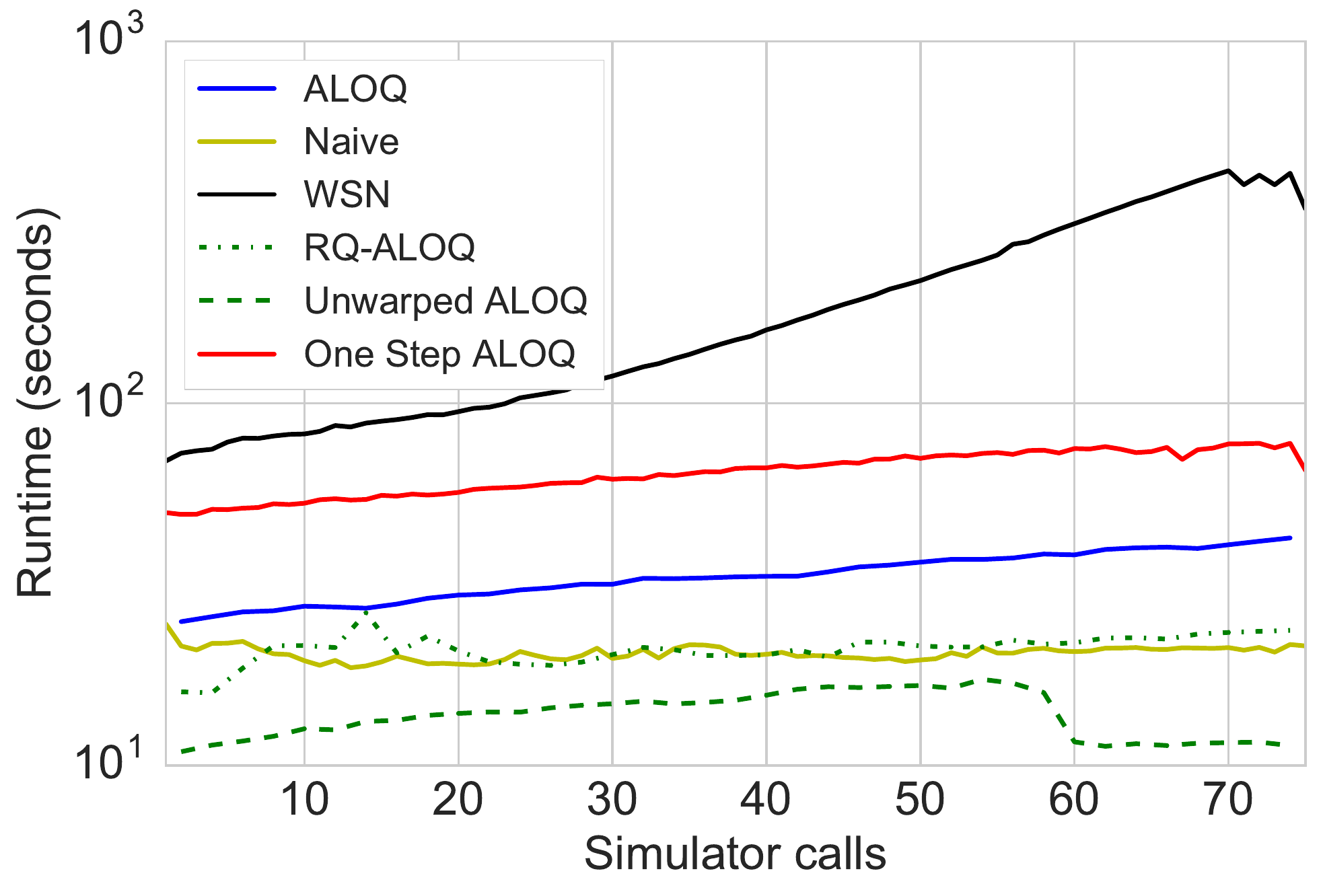}
		\caption{Branin (min)}
		\label{fig:branin_time}
	\end{subfigure}
	\begin{subfigure}{0.85\linewidth}
		\centering
		\includegraphics[width=1\linewidth]{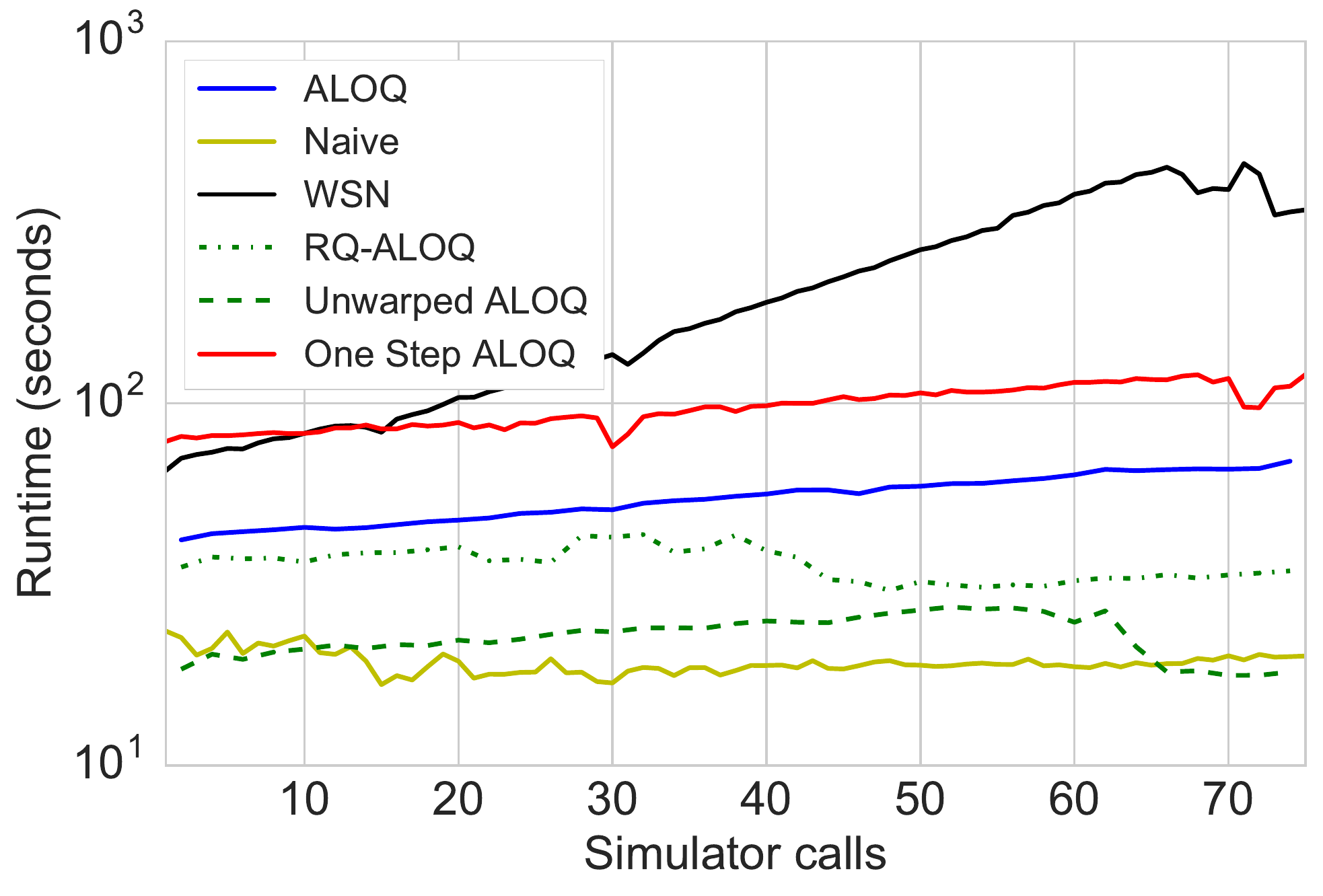}
		\caption{Hartmann 6 (max)}
		\label{fig:hart_time}
	\end{subfigure}\hfill
	\caption{Comparison of runtime of all methods on the modified Branin and Hartmann 6 test function used by \citeauthor{williams_santner}.}
	\label{fig:wsn_time_plots}
\end{figure}

Figure \ref{fig:wsn_time_plots} plots in log scale the per-step runtime of each algorithm, i.e., the time taken to process one data point on the two test functions. WSN takes significantly longer than ALOQ or the other baselines, and shows a clear increasing trend. The reduction in time near the end is a computational artefact due to resources being freed up as some runs finish faster than others.

The slow runtime of WSN is as expected due to the reasons mentioned in the paper.  However, its failure to outperform RQ-ALOQ is surprising as these are the test problems \citeauthor{williams_santner} use in their own evaluation.  However, they never compared WSN to these (or any other) baselines. Consequently, they never validated the benefit of modelling $\theta$ explicitly, much less selecting it actively. In retrospect, these results make sense because the function is not characterised by significant rare events and there is no other a priori reason to predict that simpler methods will fail.


\begin{table}[!ht]
	\caption{Quartiles of the expected function value of the final $\hat{\pi}^*$ estimated by each algorithm across 20 independent runs for each of the four artificial test functions.}
	\label{tab:test_fn}
	\begin{subtable}[h]{1\linewidth}
		\centering
		\begin{tabular}{l c  c  c}
			\toprule
			Algorithm & Q1 & Median & Q2 \\
			\midrule
			ALOQ & 326.5 & 330.0 & 352.1 \\
			Na\"{i}ve & 487.3 & 645.9 & 857.0 \\
			WSN & 519.2 & 570.8 & 735.7 \\
			RQ-ALOQ & 335.5 & 348.0 & 391.7 \\
			Unwarped ALOQ & 325.7 & 327.7 & 351.4 \\
			One Step ALOQ & 324.2 & 326.2 & 331.3 \\
			\bottomrule
		\end{tabular}
		\caption{Branin (min)}
		\label{tab:branin}
	\end{subtable}
	\begin{subtable}[h]{1\linewidth}
		\centering
		\begin{tabular}{l  c  c  c}
			\toprule
			Algorithm & Q1 & Median & Q2 \\
			\midrule
			ALOQ & 0.211 & 0.937 & 1.010 \\
			Na\"{i}ve & 0.122 & 0.823 & 1.093 \\
			WSN & 0.996 & 1.100 & 1.124 \\
			RQ-ALOQ & 0.099 & 0.214 & 0.899 \\
			Unwarped ALOQ & 0.304 & 0.306 & 1.118 \\
			One Step ALOQ & 0.210 & 1.093 &  1.118 \\
			\bottomrule
		\end{tabular}
		\caption{Hartmann 6 (max)}
		\label{tab:hart6}
	\end{subtable}
	\hfill
	\begin{subtable}[h]{1\linewidth}
		\centering
		\begin{tabular}{l c  c  c}
			\toprule
			Algorithm & Q1 & Median & Q2 \\
			\midrule
			ALOQ & 0.504 & 0.636 &  0.657 \\
			Na\"{i}ve & -0.081 & 0.081 & 0.133 \\
			WSN & -0.381 & 0.081 & 0.201 \\
			RQ-ALOQ & 0.041 & 0.081 & 0.485 \\
			Unwarped ALOQ & 0.081 &  0.523 & 0.619 \\
			One Step ALOQ & 0.596 &  0.646 & 0.655 \\
			\bottomrule
		\end{tabular}
		\caption{F-SRE1}
		\label{tab:F-SRE1}
	\end{subtable}
	\hfill
	\begin{subtable}[h]{1\linewidth}
		\centering
		\begin{tabular}{l c  c  c}
			\toprule
			Algorithm & Q1 & Median & Q2 \\
			\midrule
			ALOQ & 2.387 & 2.407 & 2.410 \\
			Na\"{i}ve & 1.917 & 1.917 & 1.917 \\
			WSN & 1.834 & 1.917 & 1.970 \\
			RQ-ALOQ & 1.916 & 1.917 & 2.295 \\
			Unwarped ALOQ & 1.908 & 2.261 & 2.410 \\
			One Step ALOQ & 2.400 &  2.405 & 2.408 \\
			\bottomrule
		\end{tabular}
		\caption{F-SRE2}
		\label{tab:F-SRE2}
	\end{subtable}
\end{table}


These results underscore the fact that a meaningful evaluation must include a problem with SREs, as such problems do demand more robust methods.  To create such an evaluation, we formulated two test functions, F-SRE1 and F-SRE2, that are characterised by significant rare events.  For $\pi \in [-2,2]$, F-SRE1 is defined as:
\begin{align}
\begin{split}
f_{F-SRE1}(\pi, \theta) =& 75\pi \exp(-\pi^2-(4\theta+2)^2)\\
&+ \sin(2\pi) \sin(2.7\theta), \\
\text{with} \hspace{0.2cm} p(\theta = \theta_j) =&
\begin{cases}
0.47\% & \text{for} \ \theta_j = -1.00, -0.95, ..., 0.00 \\
1.0\% & \text{for} \ \theta_j = 0.05, 0.10, ..., 4.50.\\
\end{cases}
\end{split}
\end{align}
And for $\pi \in [-2,2]$, F-SRE2 is defined as:
\begin{align}
\begin{split}
f_{F-SRE2}(\pi, \theta) =& \sin^2\pi + 2\cos\theta \\
&+ 200 \cos(2\pi) (0.2-\min(0.2, |\theta|)),\\
\text{with} \hspace{0.2cm} p(\theta = \theta_j) =&
\begin{cases}
1.2\% & \text{ for} \ \theta_j = -1.00, -0.98 ..., -0.22 \\
0.2\% & \text{ for} \ \theta_j = -0.20, -0.18, ..., 0.20 \\
1.2\% & \text{ for} \ \theta_j = 0.22, 0.24 ..., 1.00. \\
\end{cases}
\end{split}
\end{align}

\begin{figure}[t]
	\centering
	\begin{subfigure}{0.85\linewidth}
		\centering
		\includegraphics[width=1\linewidth]{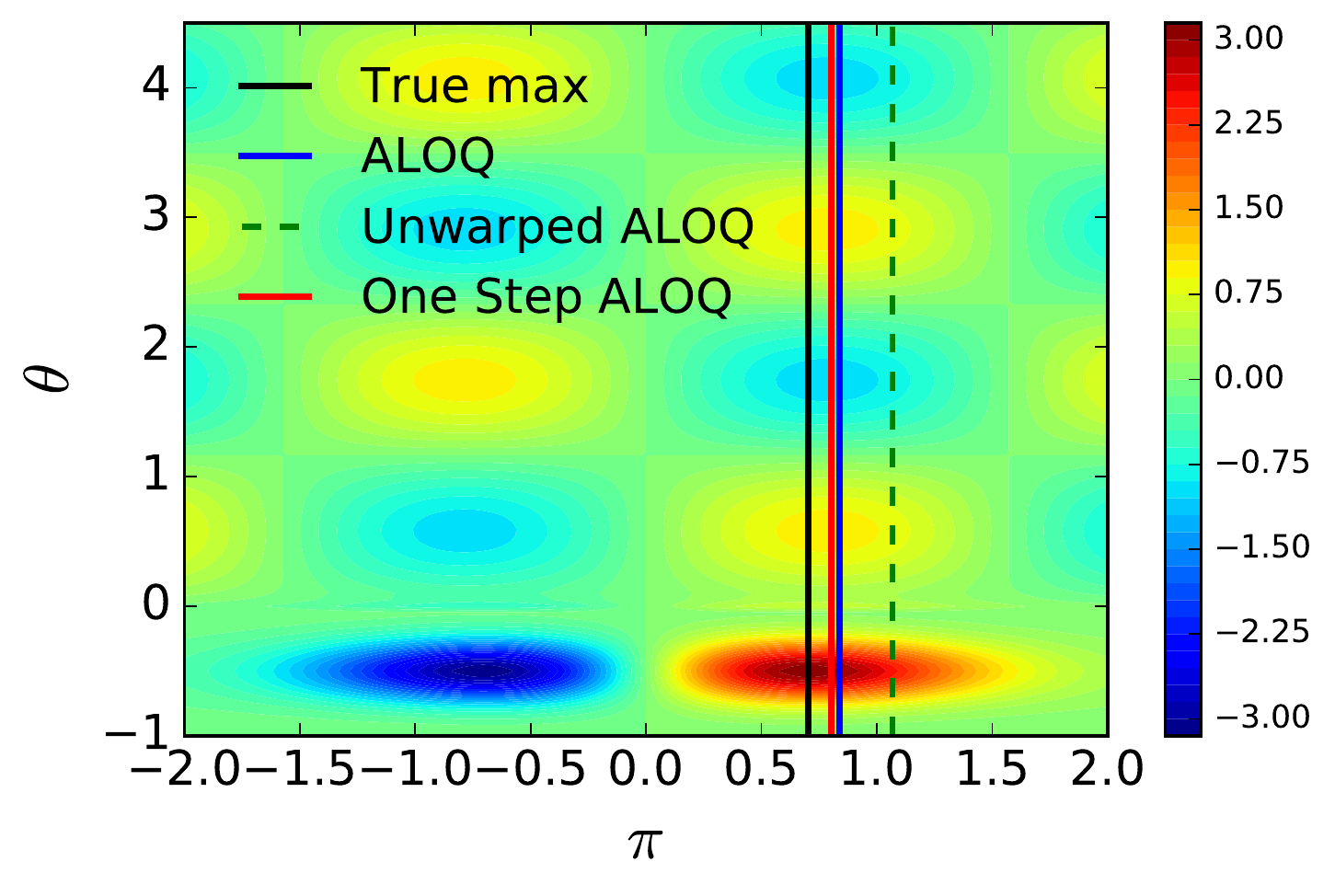}
		\caption{F-SRE1}
		\label{fig:F-SRE1_contour}
	\end{subfigure}\hfill
	\begin{subfigure}{0.85\linewidth}
		\centering
		\includegraphics[width=1\linewidth]{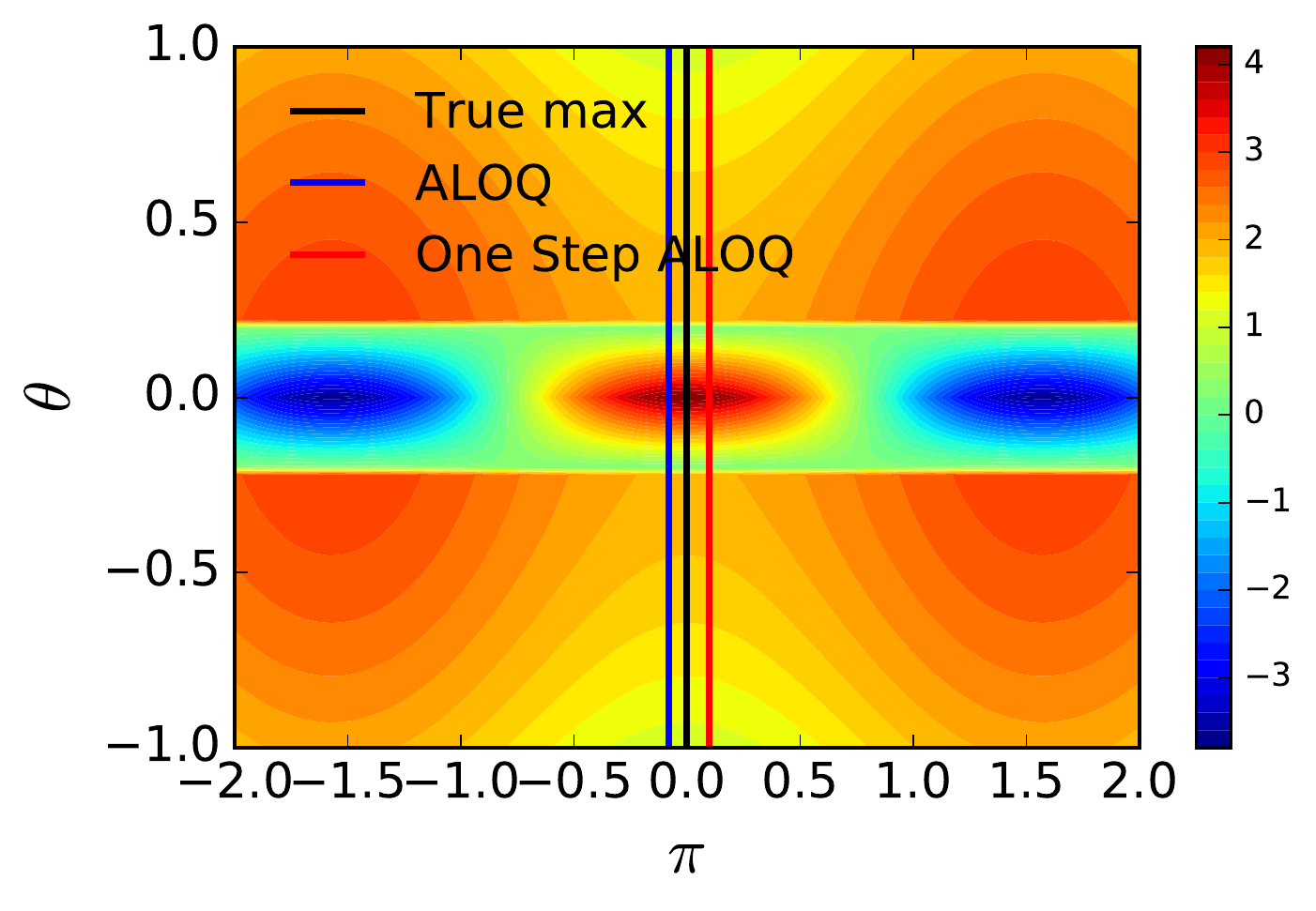}
		\caption{F-SRE2}
		\label{fig:F-SRE2_contour}
	\end{subfigure}
	\caption{Contour plot of F-SRE1 and F-SRE2 (values in SRE region have been reduced by a factor of 10).}
	\label{fig:F-SRE_contours}
\end{figure}

Figure \ref{fig:F-SRE_contours} shows the contour plots of these two functions. Both functions have a narrow band of $\theta$ which corresponds to the SRE regions, i.e. the scale of the rewards is much larger in these regions. In F-SRE1 this is $-1<\theta<0$ while in F-SRE2 this is $-0.2<\theta<0.2$. We  downscaled the region corresponding to the SRE by a factor of 10 to make the plots more readable. The final learned policy, i.e., $\hat{\pi}^*$, of each algorithm is shown as a vertical line, along with $\pi^*$ (the true maximum). These lines illustrate that properly accounting for significant rare events can lead to learning qualitatively different policies.

\begin{figure}
	\begin{subfigure}{0.85\linewidth}
		\centering
		\includegraphics[width=1\linewidth]{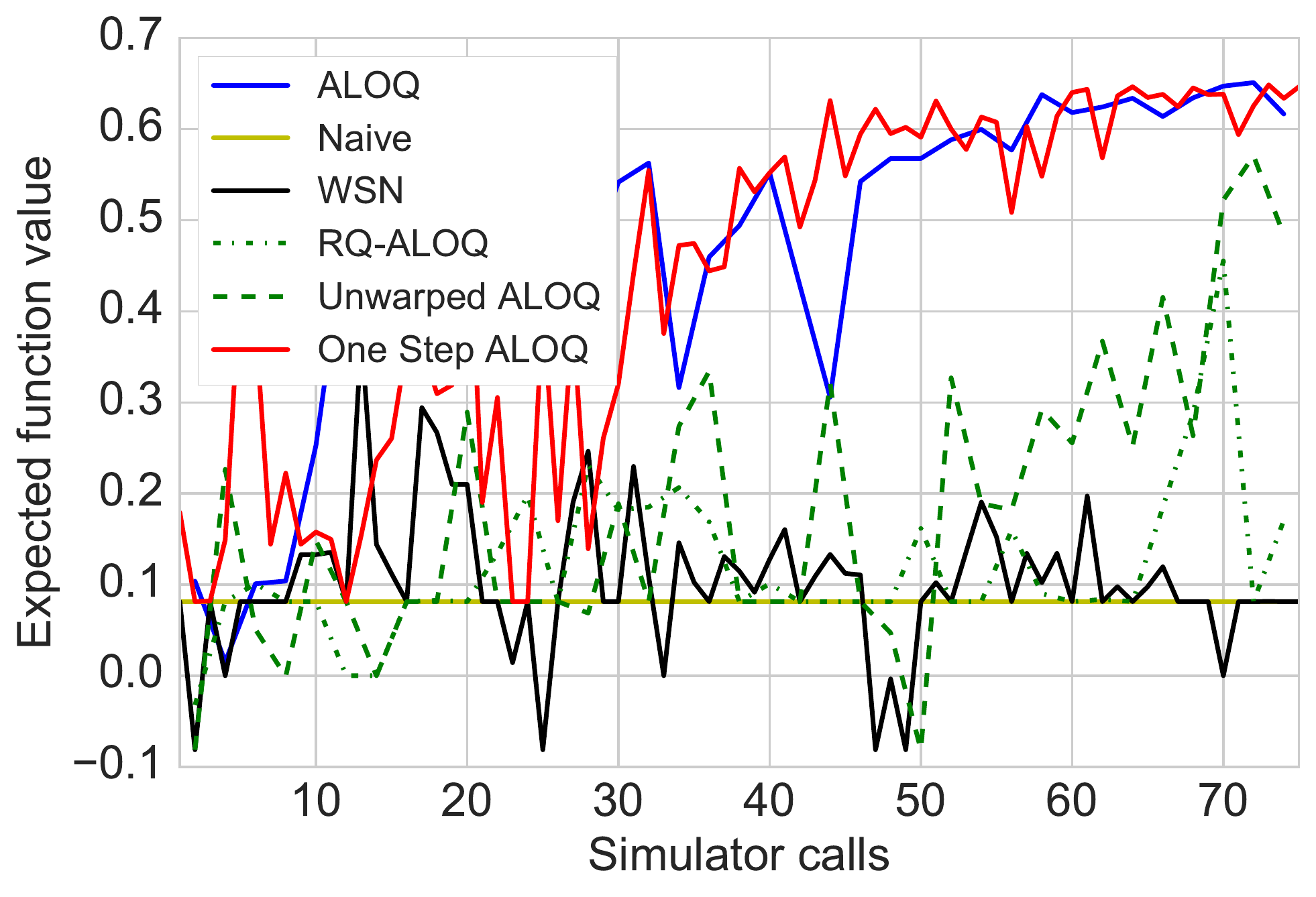}
		\caption{F-SRE1 - expected value of $\hat{\pi}^*$}
		\label{fig:F-SRE1_delta}
	\end{subfigure}\hfill
	\hskip 2em
	\begin{subfigure}{0.85\linewidth}
		\centering
		\includegraphics[width=1\linewidth]{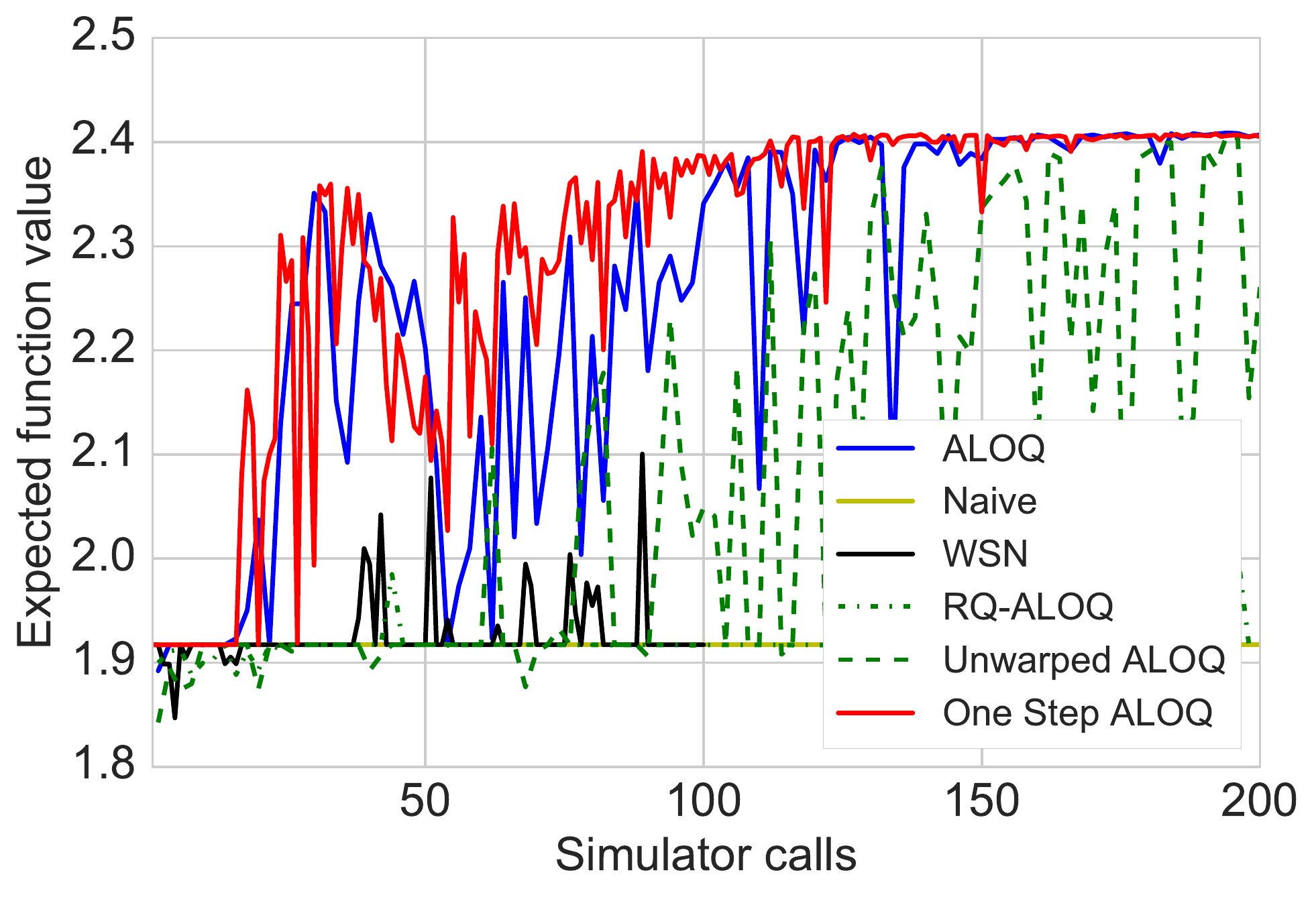}
		\caption{F-SRE2 - expected value of $\hat{\pi}^*$}
		\label{fig:F-SRE2_delta}
	\end{subfigure}
	\caption{Comparison of performance of all methods on the F-SRE test functions (higher is better)}.
	\label{fig:F-SRE_perf}
\end{figure}

Figures \ref{fig:F-SRE_perf}, which plots the performance of all methods the two functions, shows that ALOQ substantially outperforms all the other algorithms except for one-step ALOQ (note that both WSN and the na\"{i}ve approach fail completely in these settings).  As expected,  intensification does not yield any additional benefit in this low dimensional problem. However, our experiments on the robotics tasks presented in the paper show that intensification is crucial for success in  higher dimensional problems.

\begin{figure}
	\begin{subfigure}{1\linewidth}
		\centering
		\includegraphics[width=1\linewidth]{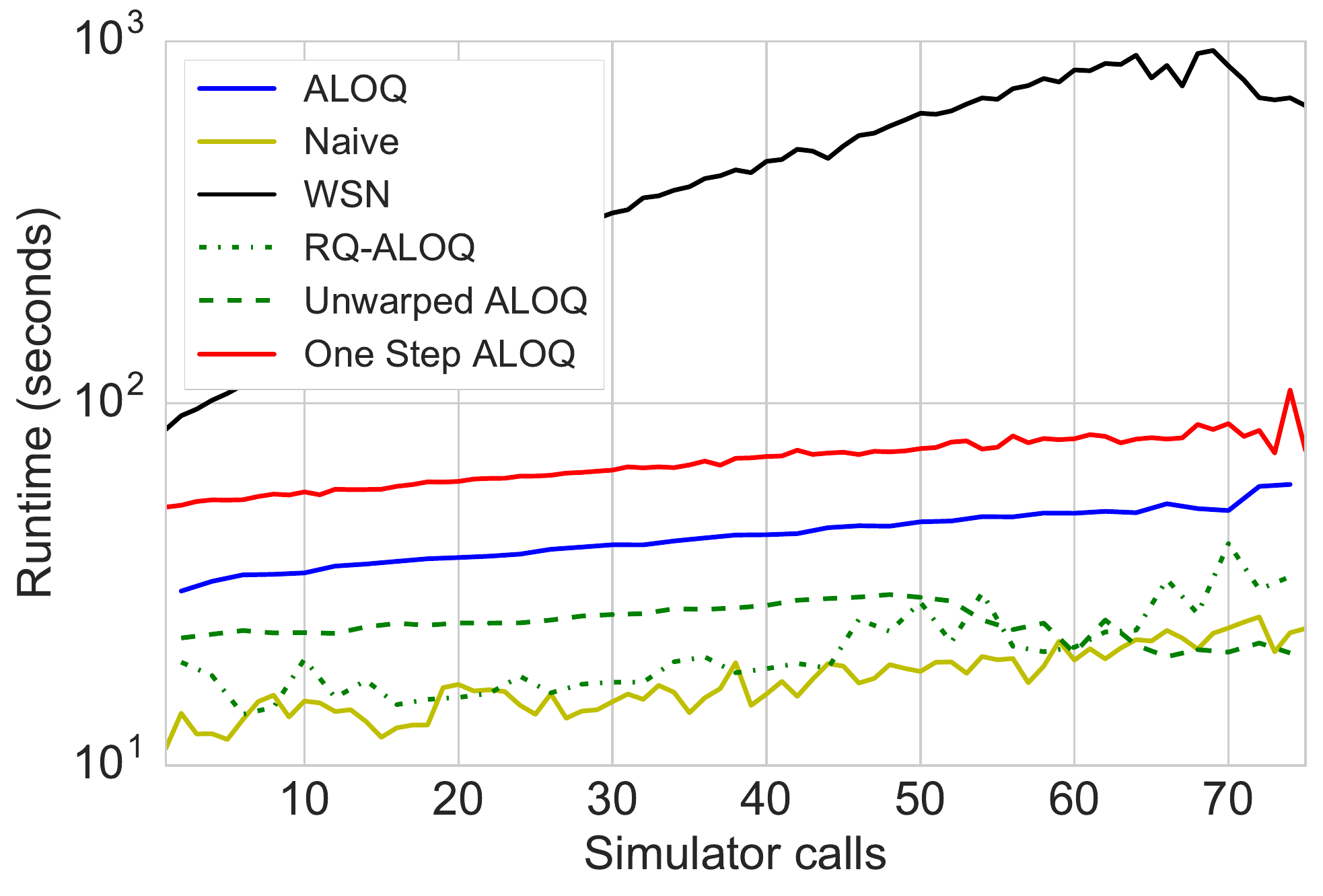}
		\caption{F-SRE1} 
		\label{fig:F-SRE1_time}
	\end{subfigure}\hfill
	\hskip 2em
	\begin{subfigure}{1\linewidth}
		\centering
		\includegraphics[width=1\linewidth]{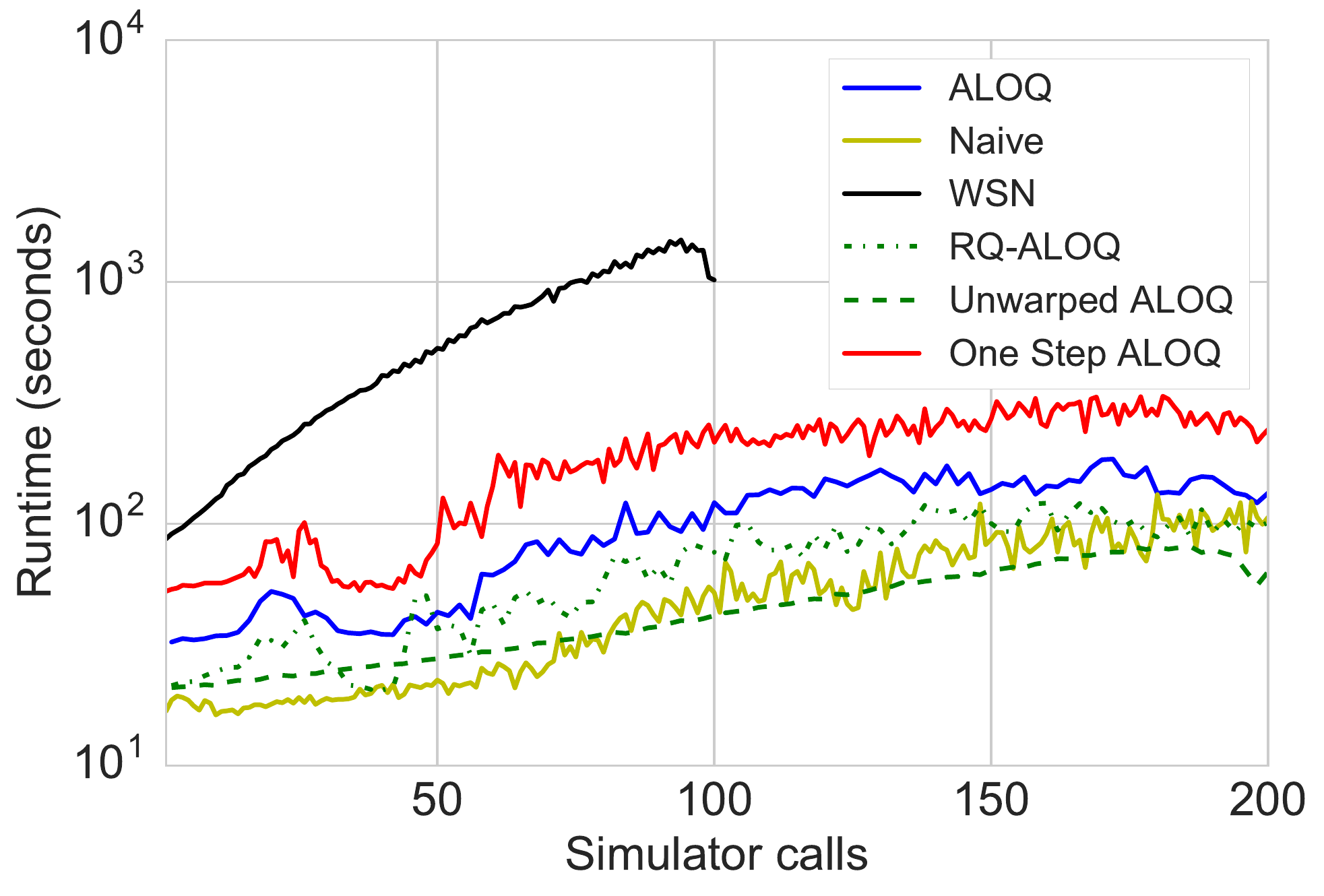}
		\caption{F-SRE2}
		\label{fig:F-SRE2_time}
	\end{subfigure}
	\caption{Comparison of runtime of all methods on the F-SRE test functions.}
	\label{fig:F-SRE_time}
\end{figure}

The per-step runtime is presented in Fig \ref{fig:F-SRE_time}. Again WSN is significantly slower than all other methods. In fact, it was not computationally feasible to run WSN beyond 100 data points for F-SRE2.

To provide a sense of the variance in the performance of each algorithm across the 20 independent runs, Table \ref{tab:test_fn} presents the quartiles of the expected function value of the final $\hat{\pi}^*$ for all four artificial test functions.

Across all four test functions, we used a log-normal hyperprior distribution with $(\mu=2, \sigma = 0.5)$ for each of $\{(\alpha_i, \beta_i)\}_{i=\pi, \theta}$ and $\kappa = 3$.

\end{document}